\documentclass[10pt,twocolumn,letterpaper]{article}

\usepackage{iccv}
\usepackage{times}
\usepackage{epsfig}
\usepackage{graphicx}
\usepackage{amsmath}
\usepackage{amssymb}

\usepackage[utf8]{inputenc}

\usepackage{bbm}
\usepackage{subcaption}

\usepackage{multirow}
\usepackage{adjustbox}
\usepackage[dvipsnames]{xcolor}

\usepackage{algorithm}
\usepackage{algorithmic}
\newcommand*{\ShowNotes}{}
\definecolor{darkred}{rgb}{0.7,0.1,0.1}
\definecolor{darkgreen}{rgb}{0.1,0.7,0.1}
\definecolor{cyan}{rgb}{0.0,0.7,0.7}
\definecolor{magenta}{rgb}{1.0,0.0,1.0}
\definecolor{dblue}{rgb}{0.2,0.2,0.8}
\definecolor{maroon}{rgb}{0.76,.13,.28}
\definecolor{burntorange}{rgb}{0.81,.33,0}

\ifdefined\ShowNotes
  \newcommand{\colornote}[3]{{\color{#1}\bf{#2: #3}\normalfont}}
\else
  \newcommand{\colornote}[3]{}
\fi

\usepackage[breaklinks=true,bookmarks=false]{hyperref}

\iccvfinalcopy % *** Uncomment this line for the final submission

\def\iccvPaperID{2484} % *** Enter the ICCV Paper ID here

\ificcvfinal\fi

\begin{document}

%%%%%%%%%%%%%%%%%% teaser
\teaser{
\begin{tabular}{c c c c c c c}
        %  \includegraphics[width=.25\linewidth] {examples/teaser/modelnet40/airplane_635/_rank_1_weight_19_seashell_400_steps.png} &
        % \includegraphics[width=.25\linewidth,]{examples/teaser/modelnet40/plant_0305/_rank_2_weight_21_noedges.png} &
        % \raisebox{0.5in}{most attended} &
        \raisebox{0.5in}{\begin{tabular}{@{}c@{}}\color{cyan}{most} \\ \color{cyan}{attentive}\end{tabular}} &
        \includegraphics[height=.18\linewidth]{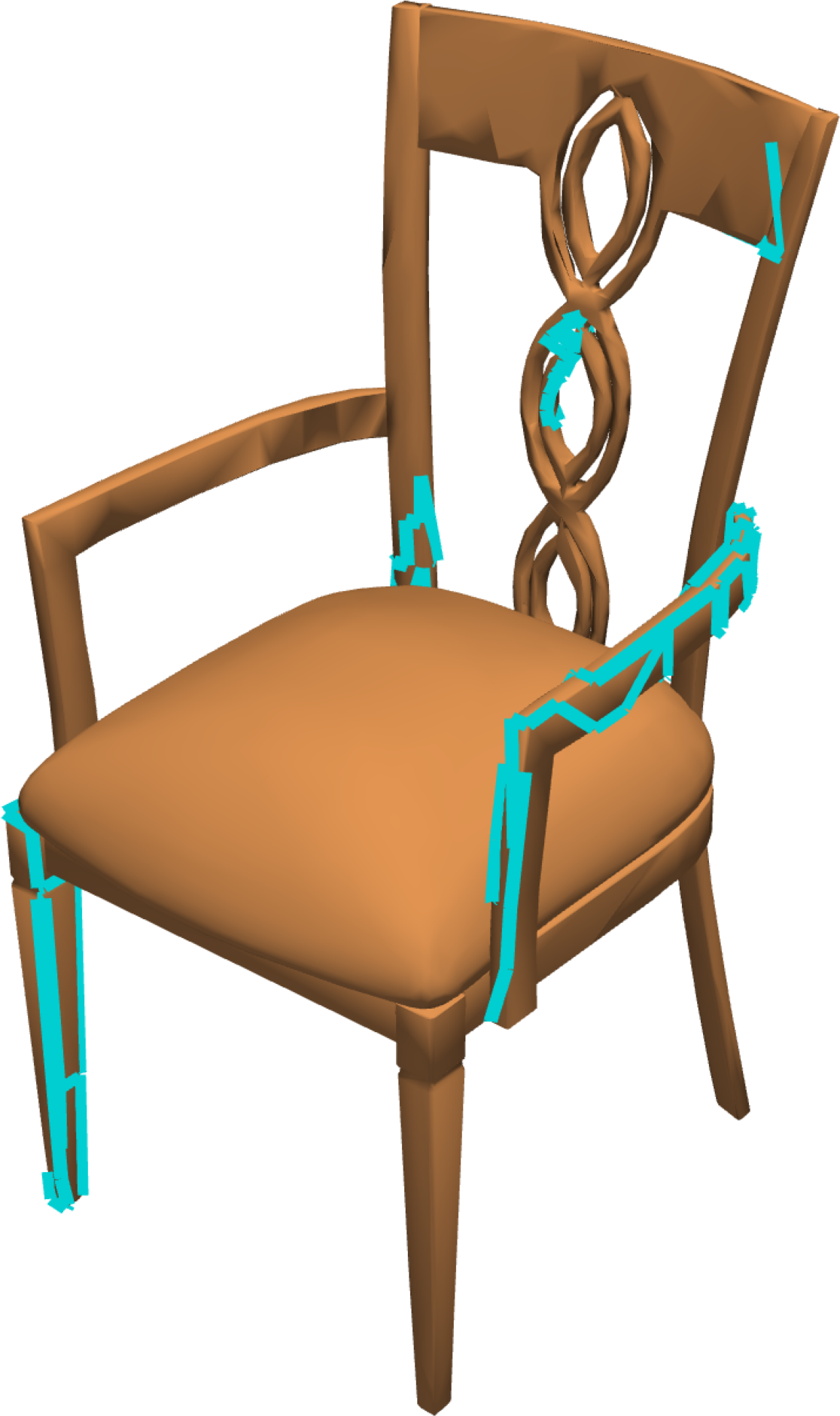} &
        \includegraphics[height=.17\linewidth]{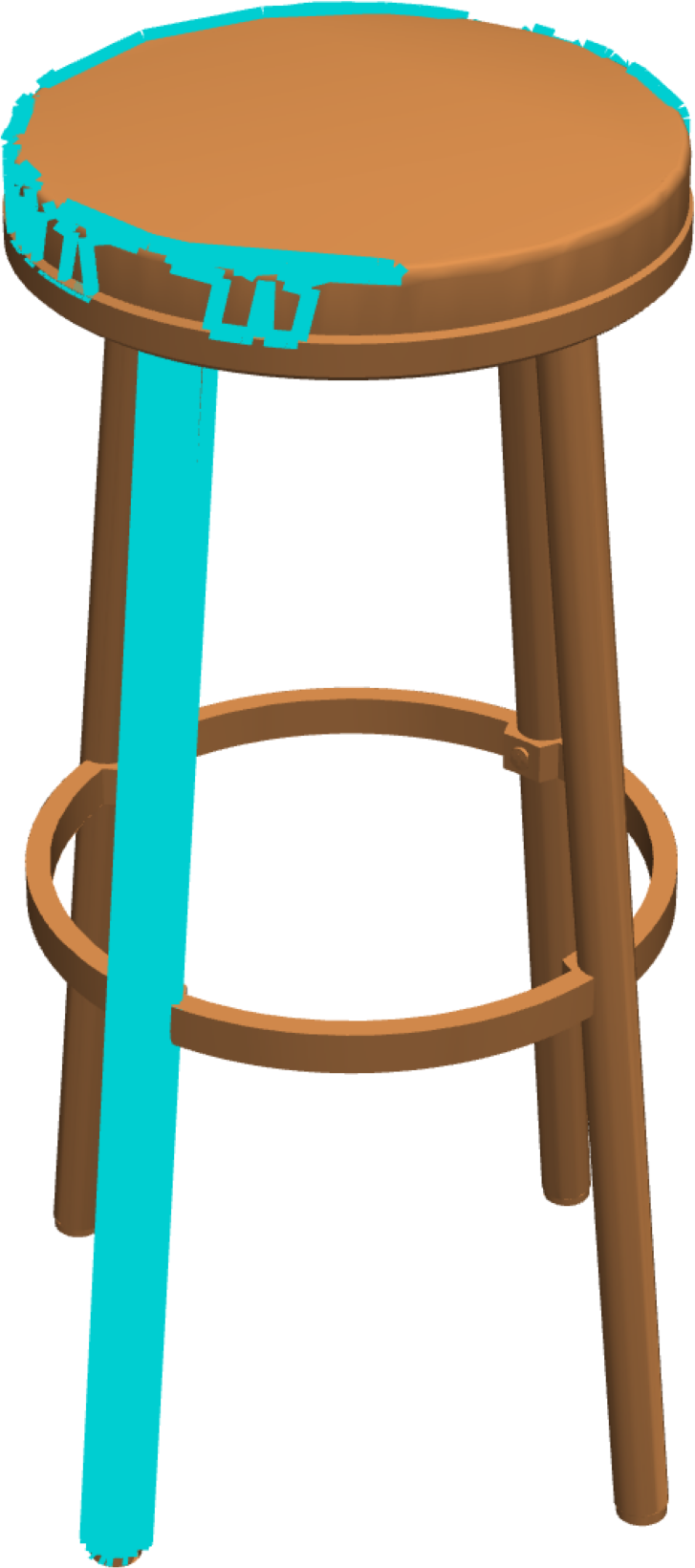} &
        \includegraphics[height=.19\linewidth]{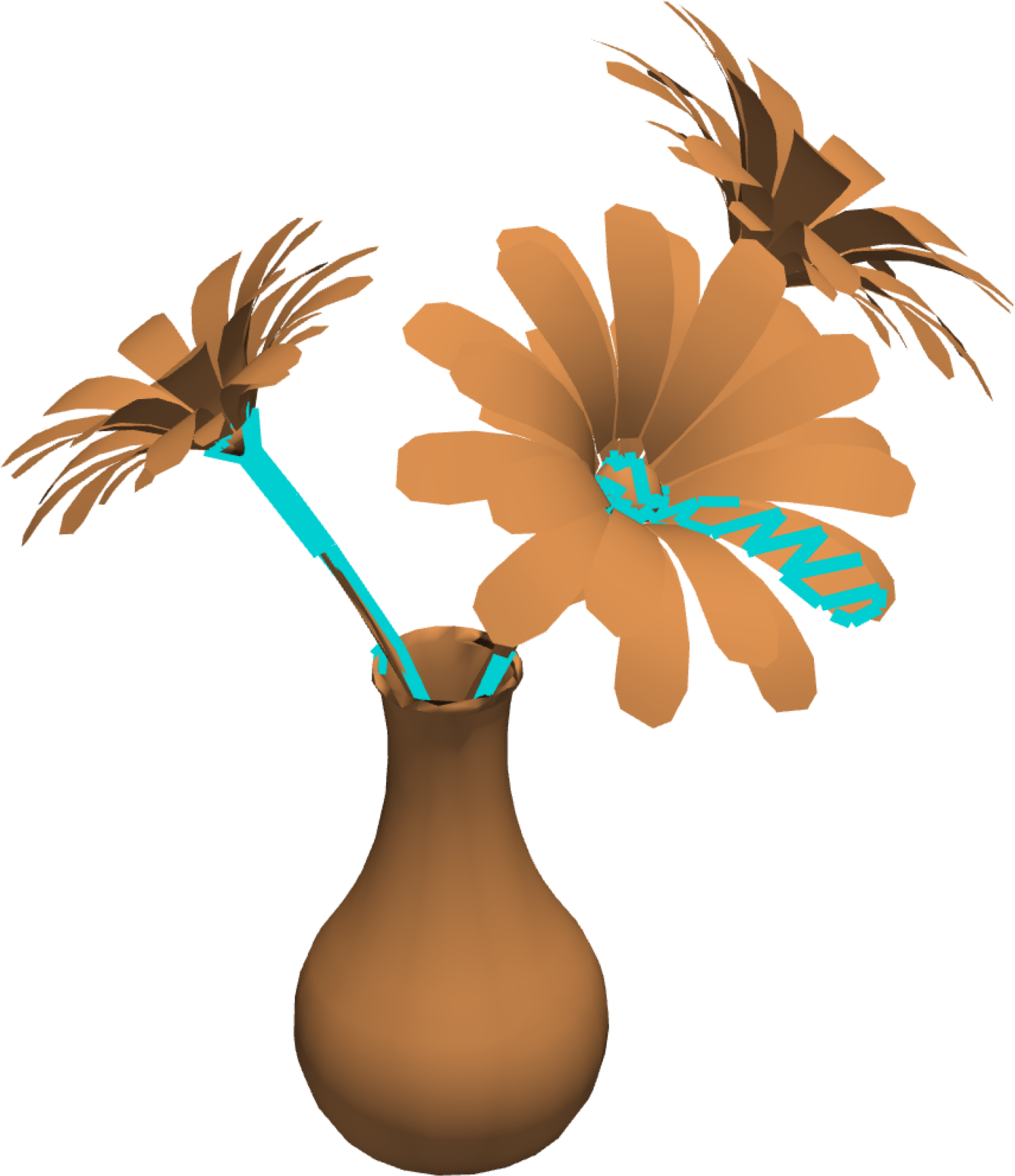}&
        \hspace{-0.15in}
        \includegraphics[height=.18\linewidth,]{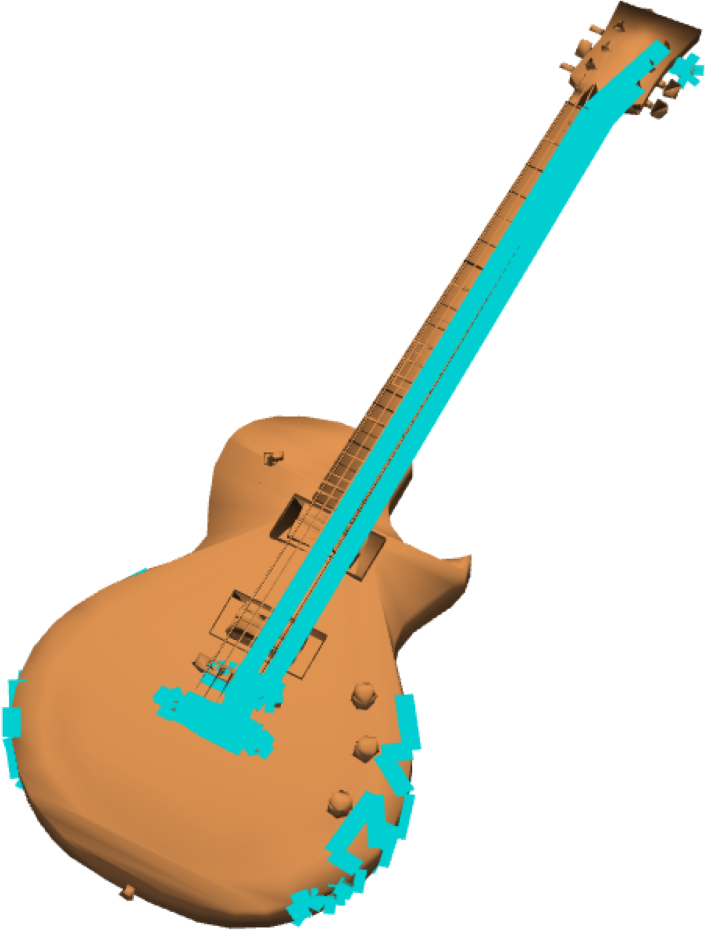} &
        \hspace{-0.05in}
        \includegraphics[height=.17\linewidth]{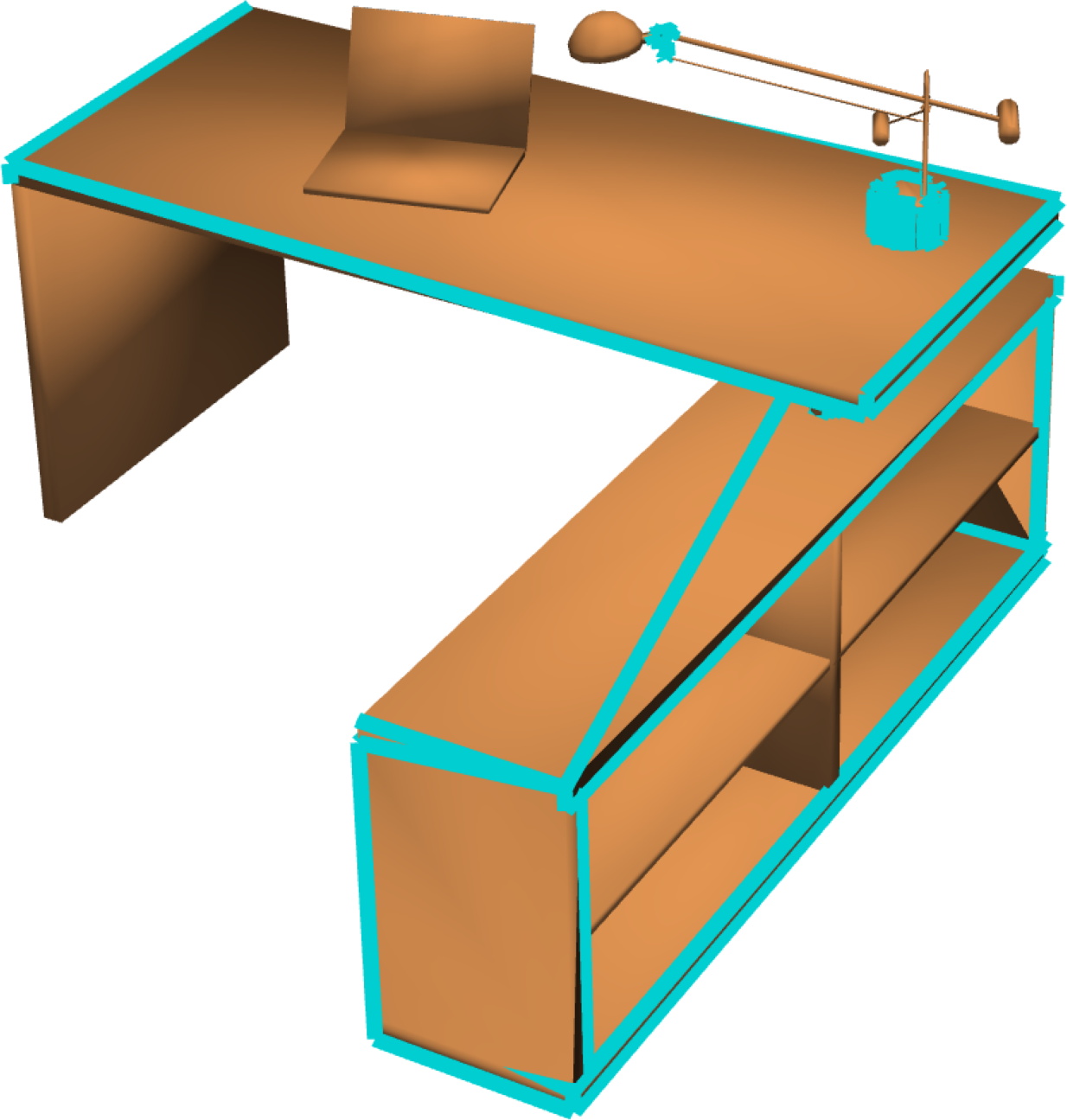} &
        \includegraphics[height=.14\linewidth]{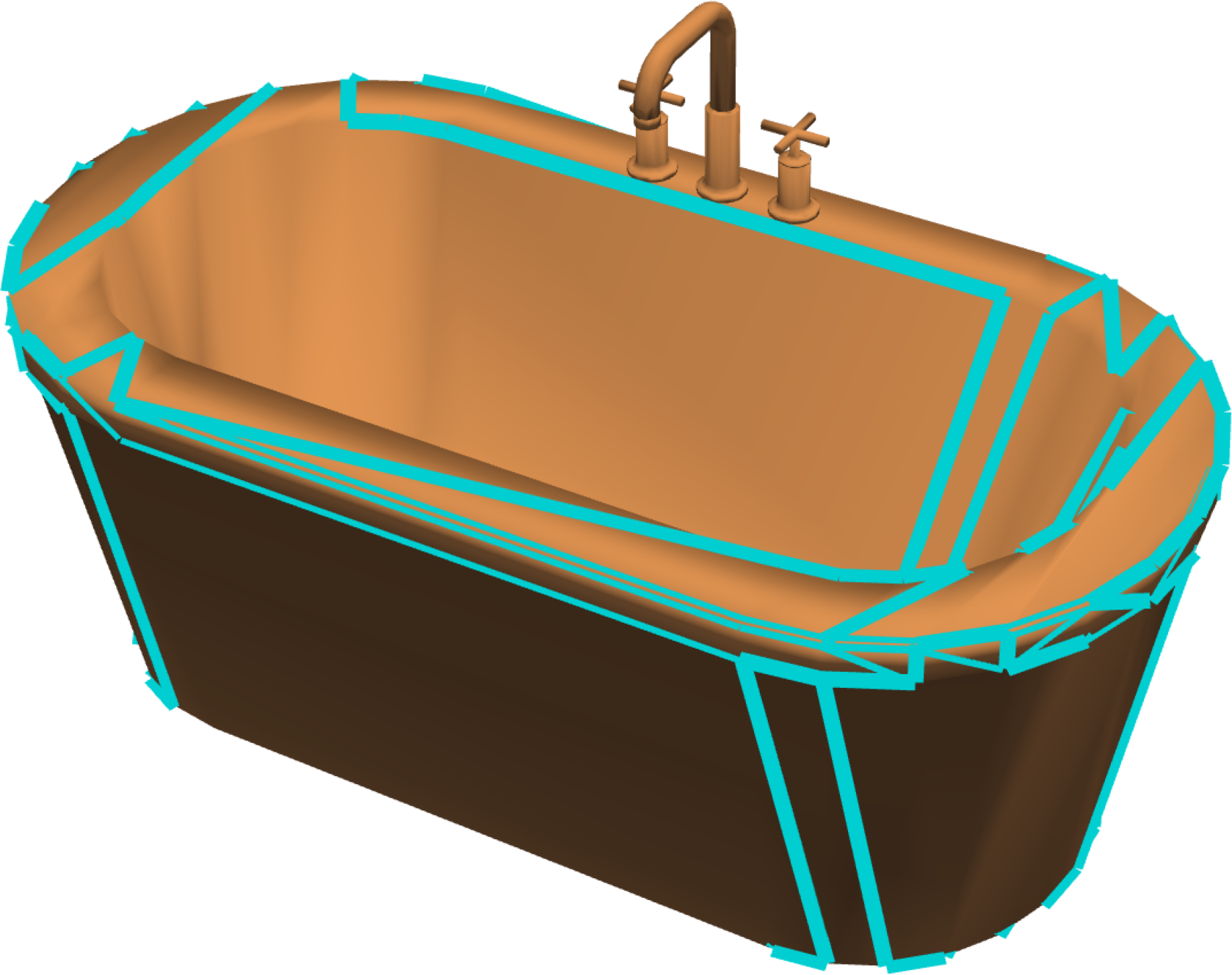}
   \\
        %  \includegraphics[width=.25\linewidth]{examples/teaser/modelnet40/airplane_635/_rank_8_weight_09_seashell_400_steps.png} &
        % \includegraphics[width=.25\linewidth]{examples/teaser/modelnet40/plant_0305/_rank_8_weight_07_noedges.png} &
        % \raisebox{0.5in}{least attentive} &
        \raisebox{0.5in}{\begin{tabular}{@{}c@{}}\color{magenta}{least} \\ \color{magenta}{attentive}\end{tabular}} &
        \includegraphics[height=.18\linewidth]{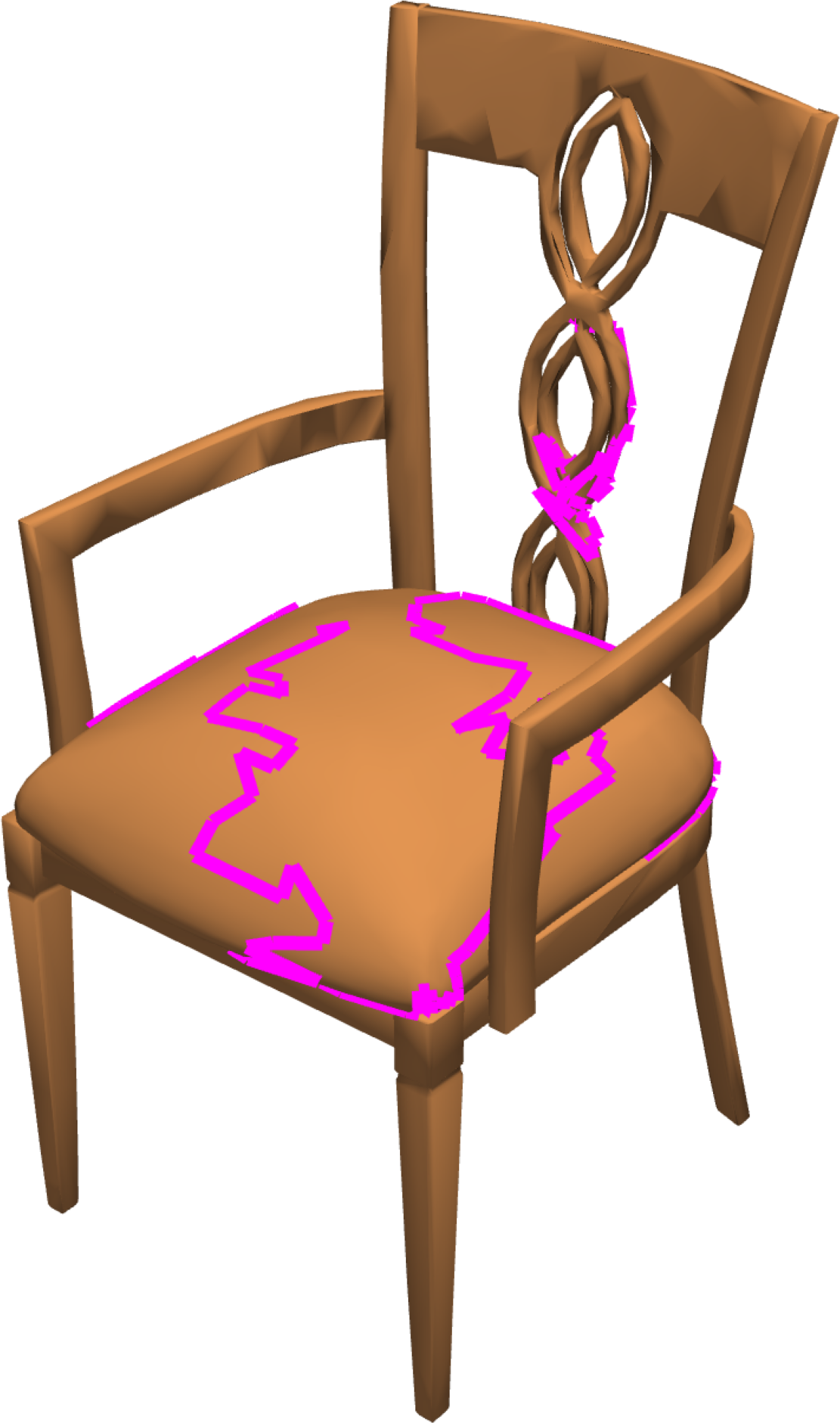} &
         \includegraphics[height=.17\linewidth]{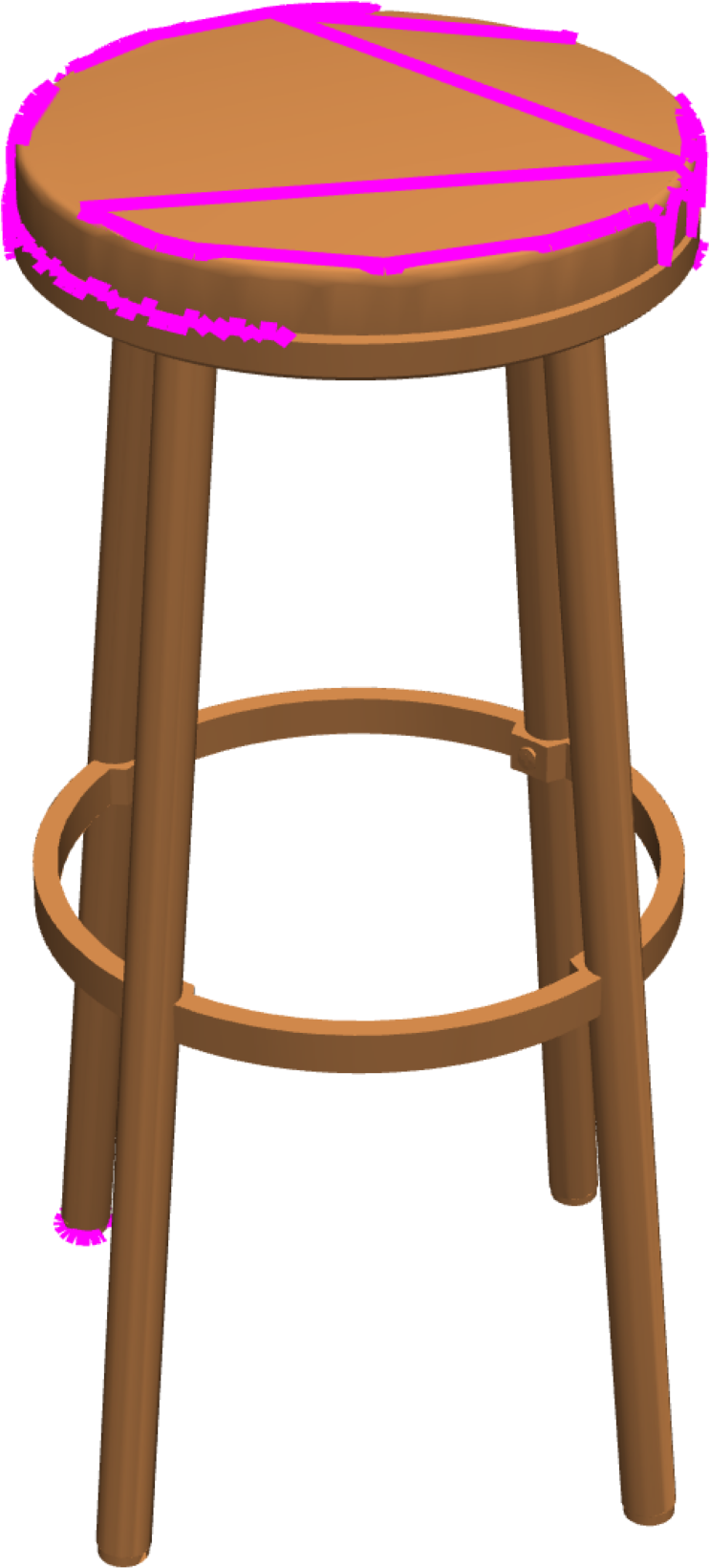} &
        \includegraphics[height=.19\linewidth]{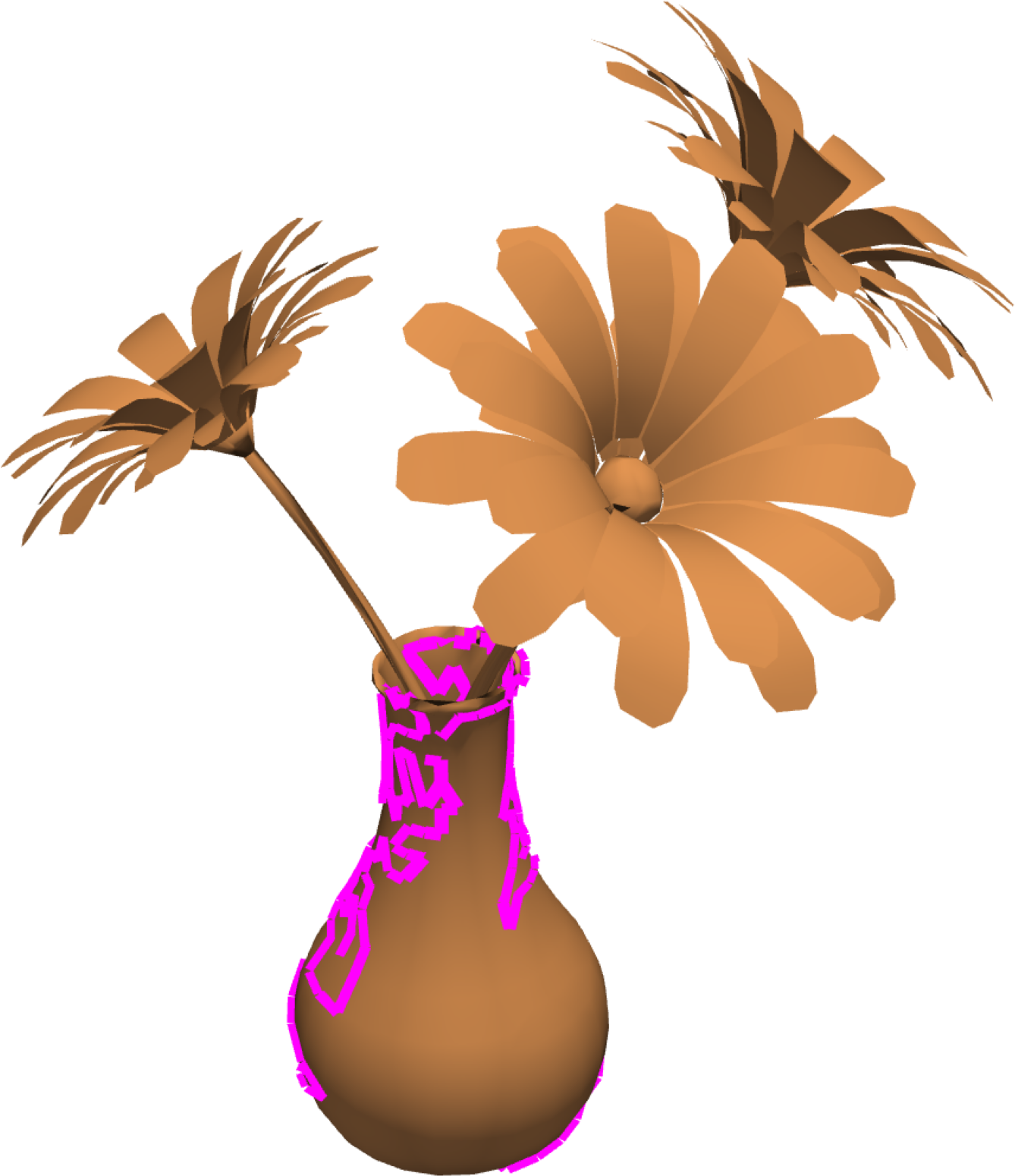}& 
        \hspace{-0.15in}
       \includegraphics[height=.18\linewidth,]{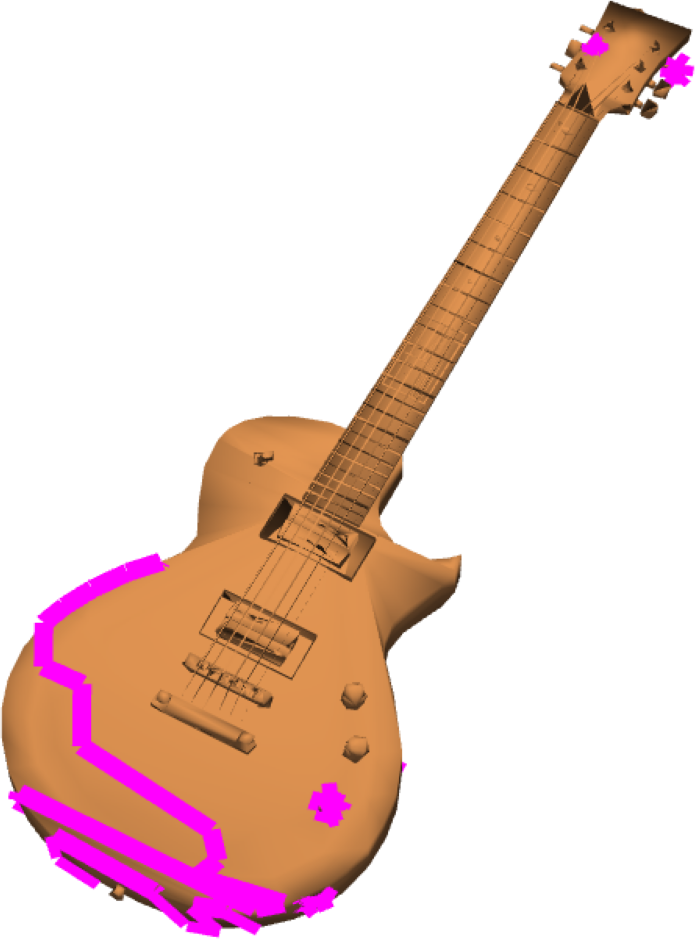} &
       \hspace{-0.05in}
        \includegraphics[height=.17\linewidth]{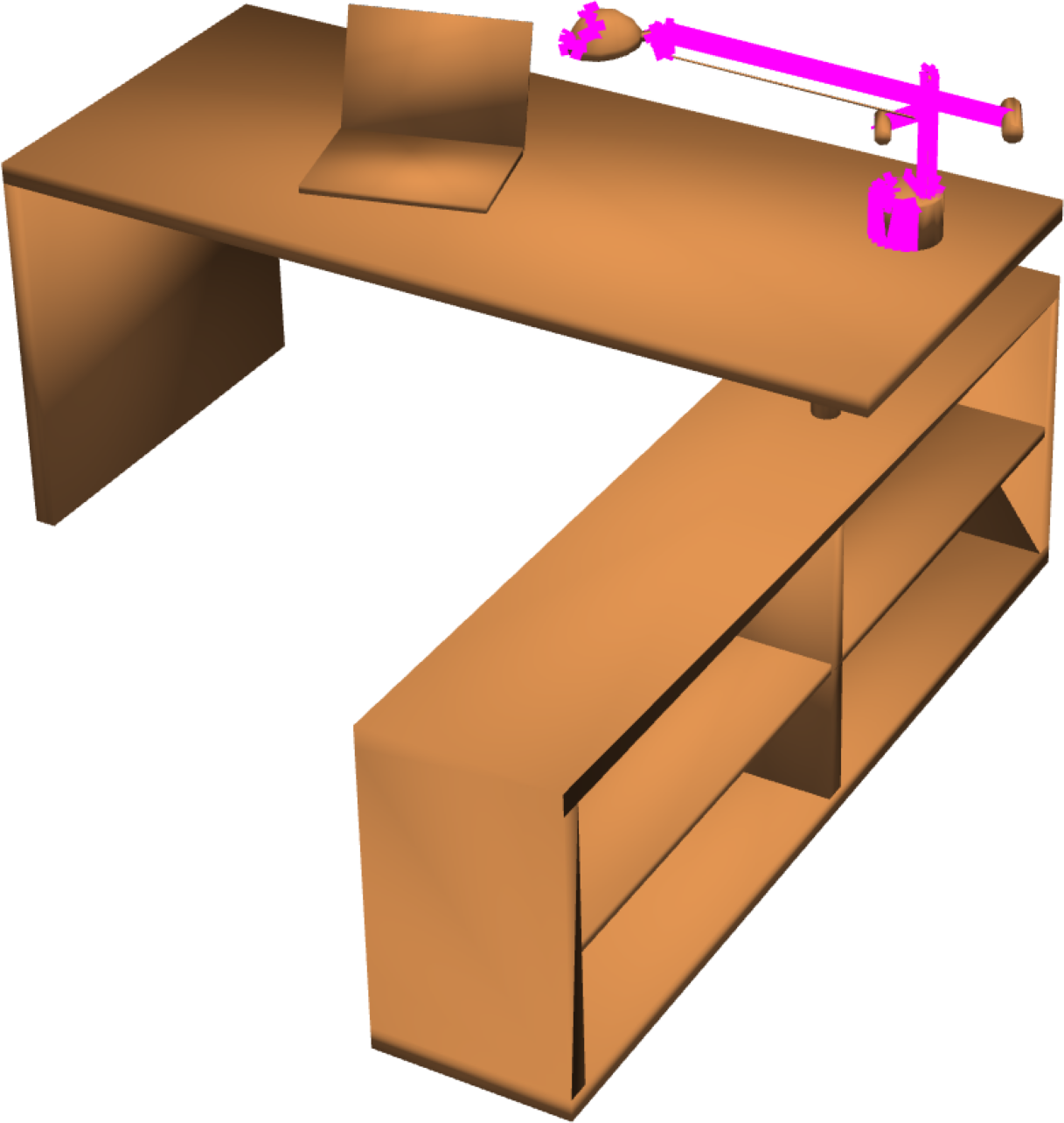} &
        \includegraphics[height=.14\linewidth]{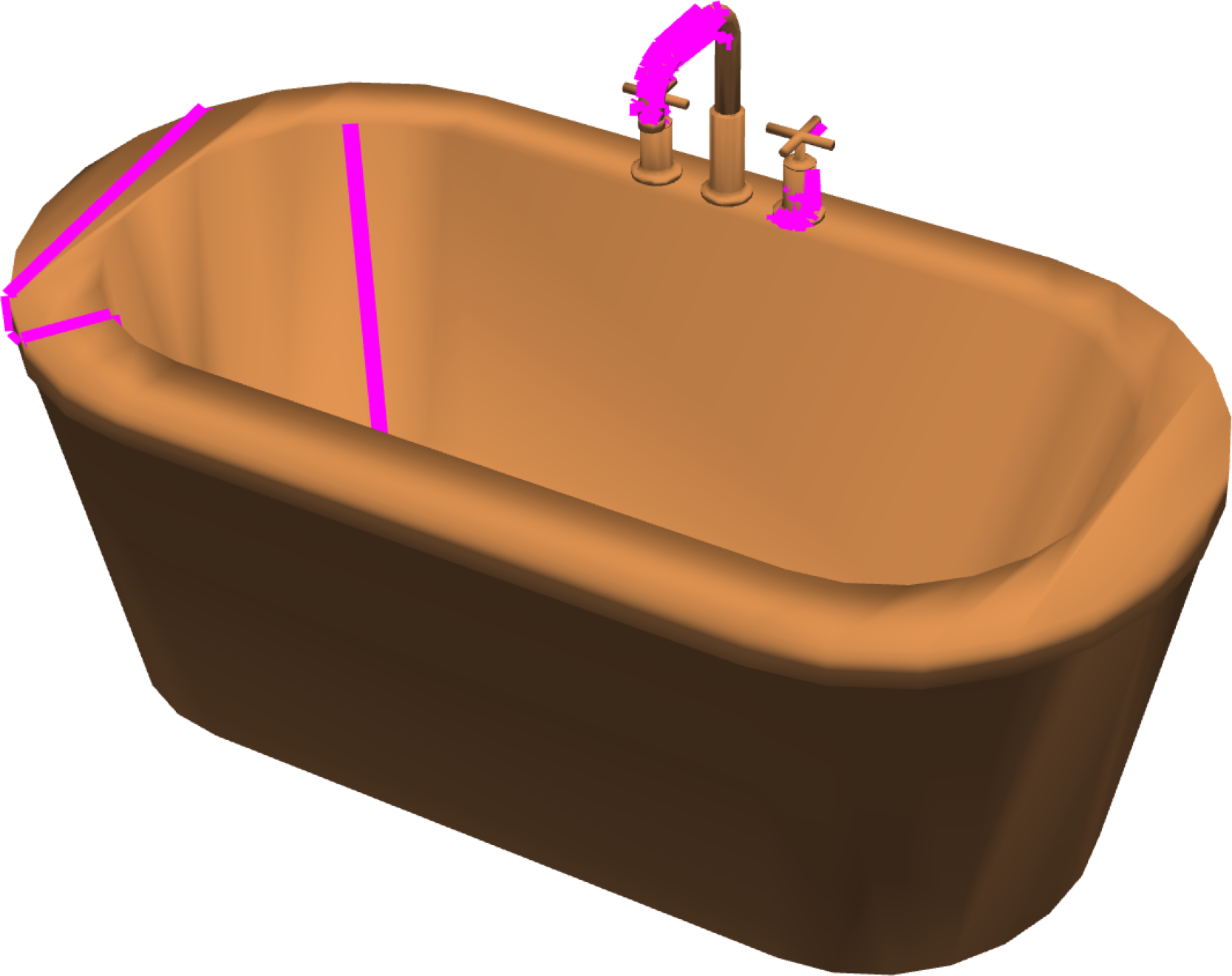}
\end{tabular}
\caption{{\bf Most \& least attentive walks.}
A set of random walks over a surface was shown to represent meshes well for deep learning.
Which walks should contribute more to the representation? 
The most attentive walks (in cyan) provide a general "view" of the object and explore its distinctive features, e.g. the guitar's neck and strings.
In contrast, the least attentive walks (in magenta) focus on regions that do not distinguish the object from others, e.g. the round seat of the stool.
}
\label{fig:teaser}
}

%%%%%%%%% TITLE
\title{AttWalk: Attentive Cross-Walks for Deep Mesh Analysis}
\author{Ran Ben Izhak\\
Technion, Israel\\
{\tt\small benizhakran@gmail.com}
\and
Alon Lahav\\
Technion, Israel\\
{\tt\small alon.lahav2@gmail.com}
\and
Ayellet Tal\\
Technion, Israel\\
{\tt\small ayellet@ee.technion.ac.il}
}

\maketitle
\ificcvfinal\thispagestyle{empty}\fi

%%%%%%%%% ABSTRACT
\begin{abstract}
Mesh representation by random walks has been shown to benefit deep learning.
Randomness is indeed a powerful concept.
However, it comes with a price---some walks might wander around non-characteristic regions of the mesh, which might be harmful to shape analysis, especially when only a few  walks are utilized.
We propose a novel walk-attention mechanism that leverages the fact that multiple walks  are used. 
The key idea is that the walks may provide each other with information regarding the meaningful (attentive) features of the mesh.
We utilize this mutual information to extract a single descriptor of the mesh. 
This differs from common attention mechanisms that use  attention to improve the representation of  each individual descriptor.
%Each walk is organized as a list of vertices, which in some manner imposes regularity on the mesh. 
%The walk is fed into a Recurrent Neural Network (RNN) that "remembers" the history of the walk. 
Our approach achieves SOTA results for two basic 3D shape analysis tasks:  classification and retrieval. 
Even a handful of walks along a mesh suffice for learning. 
\end{abstract}

\maketitle

%%%%%%%%%%%%%%%%%%%%%%%%%%%%
%%%% section: Introduction
%%%%%%%%%%%%%%%%%%%%%%%%%%%%
\section{Introduction}

% Motivation
Shape analysis of 3D objects is a fundamental aspect in modern computer vision and computer graphics research.
This is due to the paramount importance of shape analysis to numerous applications, including self-driving cars, virtual \& augmented reality, robotics, medicine and many more.

% Representations
There are several representations of 3D objects, most notably triangular meshes, point clouds and volumetric data.
This work focuses on triangular meshes, which are the most common representation in computer graphics, thanks to its efficiency and high-quality.
However, 3D meshes are unordered and irregular, which is challenging for deep learning algorithms.
This has led to attempts to "re--order" the data
%so that it would suit {\em Convolutional Neural Networks (CNNs)}. 
% These attempts include conversions to volumetric grids~\cite{ben20183dmfv, maturana2015voxnet, roynard2018classification, SZB17a} or multiple 2D projections~\cite{boulch2017unstructured, feng2018gvcnn, kanezaki2018rotationnet, su2015multi, yavartanoo2018spnet}.
% Point clouds have been handled quite intensively as well, resulting in interesting convolution and pooling operators~\cite{atzmon2018point, hua2018pointwise, li2018pointcnn, thomas2019kpconv, xu2018spidercnn}. 
% Recently, implicit functions have been proposed to represent 3D mesh for deep learning~\cite{chabra2020deep, genova2019learning, genova2020local, jiang2020local, mescheder2019occupancy, park2019deepsdf, xudisn, zakharov2020autolabeling}.
%
%  Related work 
% Our work focuses on triangular meshes, where a mesh is represented as sets of of vertices $V$, triangular faces~$F$ and Edges $E$.
and re-define the convolution  \& pooling operations, in order to be able to utilize 
%{\em Convolutional Neural Networks (CNNs)}
CNNs~\cite{feng2019meshnet,hanocka2019meshcnn,verma2018feastnet,wiersma2020cnns}.
% These  include
% FeaStNet~\cite{verma2018feastnet}, 
% %proposes a graph neural network in which the neighborhood of each vertex for the convolution operation is calculated dynamically based on its features. 
% MeshCNN~\cite{hanocka2019meshcnn},
% %defines pooling and convolution layers over the mesh edges. 
% MeshNet~\cite{feng2019meshnet}, 
% %treats the faces of a mesh as the basic unit and extracts their spatial and structural features individually to offer the final semantic representation.
%  LRF-Conv~\cite{yang2021continuous}, and
% % learns descriptors directly from the raw mesh by defining new continuous convolution kernels that provide robustness to sampling. 
% HSN~\cite{wiersma2020cnns}.  

% \cite{gao2020pairwise} - multiview work, unknown conference( SIGIR?), 3 citations (looks like self-citations, not sure), doing only retrieval.

% Alon's work in the nutshell
Our work is inspired by a recent framework, named {\em Meshwalker}~\cite{lahav2020meshwalker}, which does not attempt to use CNNs at all.
Instead, it suggests to capture the geometry \& topology of a mesh by randomly walking along its surface.
% through the vertices of the mesh, along its edges.
Each walk is processed by a {\em Recurrent Neural Network (RNN)}, aggregating surface information along the walk. 
% For each step they produce a feature vector by relating to the global (aggregated information from previous steps) and local (current vertex input) geometry of the surface.
Given a mesh, it is represented by several random walks, which are generated independently.
Meshwalker achieves outstanding results, demonstrating the power of randomness.

% key idea
In this paper, we attempt to overcome the major drawback of Meshwalker:
Randomness may result in walks that do not represent the mesh well, sometimes leading to failures even when many walks are used.
%Some walks go through meaningful regions of the shape--those that distinguish it from other shapes, while others do not.
% Mesh analysis, however, mostly regards the former.
Our goal is to focus on portions of the walk that distinguish a mesh from others.
We propose to learn how to weigh the features of the various walks.
Interestingly, this yields insight regarding the distinctiveness of mesh regions.
%W e propose to learn to identify meaningful walks---those that go through distinguish regions.
% We weigh the distinctive features of walks accordingly.
% For instance, as the distinctive potions of a guitar are its neck and strings, we would like to focus on walks that go through these parts, as can be seen in Figure~\ref{fig:teaser}.
For instance, walk features related to the guitar's neck and strings are more meaninful than those related to the guitar's body (Figure~\ref{fig:teaser}).

% attention 
This idea is related to the popular notion of attention, first introduced  by~\cite{bahdanau2015neural} for language translation.
Since then, many natural language processing  {\em (NLP)} tasks, as well as image analysis tasks have used attention.
% Since then, many {\em natural language processing (NLP)} tasks \cite{devlin2019bert, vaswani2017attention,yang2019xlnet}, as well as image analysis tasks~\cite{xu2015show,yang2016stacked,zhang2019self, zhao2020exploring} have used attention.
% ~\cite{chen2016attention, fu2019dual, wang2017residual, xu2015show,yang2016stacked,zhang2019self, zhao2020exploring}
The work on 3D attention, in particular on meshes, is sparser~\cite{lin2020end, Milano20NeurIPS-PDMeshNet, yang2020multiscale}.

% The idea idea
We propose a novel walk-attention mechanism.
We exploit the fact that multiple walks along a mesh result in multiple representations of this mesh.
These walks can provide information to each other on the important features of the walk and jointly derive a good description of the mesh.
For intuition sake, let us draw an analogy to sentences.
Suppose that we are given different sentences describing the same event (i.e., various walks describing the same mesh).
%Each sentence is processed sequentially and a feature vector is built for it, describing in some manner the event, as expressed by this sentence.
Our goal is come up with a single description of the significant features of the event, by utilizing the collection of sentences.
This differs from common attention units, which aim at improving the descriptors of the various sentence.
% Note that this analogy is not complete, as in sentences no randomness is involved.

% Realization
To realize this idea, we learn a novel attention map between the feature vectors  representing the walks  along the mesh.
Rather than using the attention vectors as new feature vectors, as common, we interpret them as cross-walk probabilities.
This will enable us to use them, jointly with the original feature vectors, to generate a single mesh descriptor.
% The higher the probability (per vector entry) of a certain walk, the more weight this walk gets in the final mesh descriptor.
This descriptor focuses on the distinctive mesh features encountered by specific random walks, while neglecting the effect of commonplace features encountered by others.

Figure~\ref{fig:teaser} illustrates the most attentive walks----those that influence the final mesh descriptor the most  (and similarly, the least attentive walks).
Indeed, the most attentive walk of the stool, as determined by our model, is one that goes both through the seat and a leg, whereas the least attentive walk goes only through the seat, and thus missing information needed for classification.
Similarly, the most attentive walk of the guitar explores the neck and the strings, and that of plant explores the flowers (neglecting the walk on the vase).

% other benefits
In addition to achieving SOTA results, our framework is able to analyze shapes using significantly less walks for certain datasets.
For instance, for Modelnet40, $\frac{1}{8}^{th}$ of the walks suffice, compared to~\cite{lahav2020meshwalker}, while the results improve.
This is thanks to focusing on the most distinctive portions of the walks, rather than averaging all random walks.

% Applications
We evaluate our method for two fundamental shape analysis applications: mesh classification and mesh retrieval. 
We show that it outperforms other methods for commonly-used datasets, as well as for a challenging new dataset~\cite{fu20203d}.
%
% Contributions
Hence, this paper makes a couple of contributions:
\begin{enumerate}
\vspace{-0.05in}
    \item 
    We introduce a novel attention mechanism to deep learning on meshes. 
%    This attention gives more weight to features of distinctive walks.
    This mechanism also provides insight as to which regions of 3D objects are more important for shape analysis tasks and which are less.
\vspace{-0.05in}
    \item
    We present an end-to-end learning framework that realizes this attention.
%    \item
    It achieves state-of-the-art results for 3D shape classification and  retrieval, even when using significantly few walks.
\end{enumerate}

%%%%%%%%%%%%%%%%%%%%%%%%%%%%
%%%% Section: Related work
%%%%%%%%%%%%%%%%%%%%%%%%%%%%
\section{Related work}
\label{sec:related}

%%%%%%%%%%%%%%% Meshes in deep learning
\noindent
{\bf Mesh deep learning.}
A triangular mesh is the most widespread 3D representation in computer graphics.
A mesh is represented as a set of vertices $\mathcal{V}$, edges $\mathcal{E}$ and faces~$\mathcal{F}$.
Since each vertex has a different number of neighbors, at different distances, the basic question is how the irregular nature of this representation shall be handled, so as to suit {\em Convolutional Neural Networks (CNNs)}.

In an attempt to "re--order" the data,
it was suggested to convert the mesh into volumetric grids~\cite{ben20183dmfv, fanelli2011real} 
% \emph{Volumetric grids.}
% These grids are the extension to 3D of images.
% Thus, processing these grids can be done by direct extension of 2D processing~\cite{brock2016generative, fanelli2011real, maturana2015voxnet, sedaghat2016orientation, tchapmi2017segcloud, wang2019normalnet, wu20153d, zhi2018toward, SZB17a}.
% Volumetric grids require heavy computation and are thus have limited resolution.
%
or into multiple 2D projections (multi-view)~\cite{feng2018gvcnn, kanezaki2018rotationnet, su2015multi, wei2020view}.
% ~\cite{boulch2017unstructured, wang2019dominant, feng2018gvcnn, kanezaki2018rotationnet, su2015multi, wei2020view, yavartanoo2018spnet}
% \emph{Multi-view 2D projections.}
% Here, a set of 2D images, each taken from a different viewpoint, represents the shape~\cite{bai2016gift, feng2018gvcnn, gomez2017lonchanet, han20193d2seqviews, he2018triplet, johns2016pairwise, kalogerakis20173d, kanezaki2018rotationnet, qi2016volumetric, sarkar2018learning, su2015multi, wang2019dominant, zanuttigh2017deep}.
% Commonly-used 2D frameworks can be utilized for this representation, which is highly beneficial.
% However, it is limited to certain applications (e.g. it cannot be used for segmentation) and to certain objects (single objects with limited self occlusion).
%
Point clouds have been handled quite intensively as well, resulting in interesting convolution and pooling operators~\cite{atzmon2018point, qi2017pointnet, qi2017pointnet++, thomas2019kpconv, xu2018spidercnn}. 
% ~\cite{atzmon2018point, li2018pointcnn, qi2017pointnet, qi2017pointnet++, thomas2019kpconv, xu2018spidercnn}. 
% \emph{Point clouds.}
% Many recent works have proposed successful techniques for point cloud shape analysis using neural networks~\cite{atzmon2018point,li2018pointcnn, liu2019relation, qi2017pointnet, qi2017pointnet++, wang2019dominant, xu2018spidercnn, zhu2019random, williams2019deep, guerrero2018pcpnet}.
% This representation consists of a set of 3D points, sampled from the object’s surface.
% This is a simple representation, which has close relationship to data acquisition, and thus it is very attractive.
% It is less useful when the connectivity is meaningful.
%
Recently, implicit functions have also been proposed to represent 3D mesh for deep learning~\cite{genova2020local, jiang2020local, mescheder2019occupancy, park2019deepsdf}.
% ~\cite{chabra2020deep, genova2020local, genova2019learning, jiang2020local, mescheder2019occupancy, park2019deepsdf, xudisn, zakharov2020autolabeling}.
See~\cite{gezawa2020review} for a thorough review.

To handle meshes directly, 
novel convolutions and/or vertex neighborhoods have been defined~\cite{feng2019meshnet, gong2019spiralnet++,poulenard2018multi,schult2020dualconvmesh,verma2018feastnet}. 
% The pioneering work of~\cite{masci2015geodesic} introduces deep learning of local features and shows how to make the convolution operations intrinsic to the mesh.
%In~\cite{poulenard2018multi} a new convolutional layer is defined, which allows the propagation of geodesic information throughout the network layers.
% FeaStNet~\cite{verma2018feastnet} proposes a graph neural network in which the neighborhood of each vertex for the convolution operation is calculated dynamically based on its features.
Other works 
% exploit the fact that local patches are approximately Euclidean.
%The 3D manifolds are then parameterized in 2D, where standard CNNs are utilized
parameterize the mesh in 2D~\cite{boscaini2016learning, ezuz2017gwcnn, henaff2015deep, maron2017convolutional, sinha2016deep}.
% ~\cite{boscaini2016learning, ezuz2017gwcnn, haim2019surface, henaff2015deep, maron2017convolutional, sinha2016deep}.
%A different approach is to apply a linear map to a spiral of neighbors~\cite{lim2018simple, gong2019spiralnet++}, which works well for meshes with a similar graph structure.
In~\cite{hanocka2019meshcnn}, a unique idea of using the edges of the mesh to perform pooling and convolution, is introduced. 

% Working directly on meshes %
% MeshNet~\cite{feng2019meshnet} treats faces of a mesh as the basic unit and extracts their spatial and structural features individually, to offer the final semantic representation.
%The convolution operations exploit the regularity of edges—having 4 edges of their incidental triangles. An edge collapse operation is used for pooling, which maintains surface topology and generates new mesh connectivity for further convolutions.
% DualConvMesh-Net~\cite{schult2020dualconvmesh} suggest applying convolution on mesh vertices both in geodesic space using graph convolutions and in euclidean space using point cloud convolutions.

Our work is based on yet another idea, termed {\em MeshWalker}, of representing a mesh by a set of random walks over its vertices, along the edges~\cite{lahav2020meshwalker}.
In a walk, each vertex is represented as its 3D offset from the previous vertex of the walk.
% Each vertex along the walk is represented as the 3D translation $(\Delta X, \Delta Y, \Delta Z)$ from the previous vertex. 
% The first vertex is uniformly sampled from the mesh vertices, and is not part of the input sequence.
The walk is fed into a {\em Recurrent Neural Network (RNN)} that "remembers" the walk's history.

%%%%%%%%%%%%%%%%%%%%% Attention
\vspace{0.05in}
\noindent
{\bf 3D attention.}
% NLP
Attention was first introduced  by~\cite{bahdanau2015neural} for language translation and since then have revolutionized {\em natural language processing (NLP)}~\cite{cheng2016long}.
% Among the NLP tasks that have have used attention are 
% translation~\cite{devlin2019bert, vaswani2017attention}, 
% inference~\cite{yang2019xlnet}, 
% %inference~\cite{parikh2016decomposable,yang2019xlnet}, 
% text classification~\cite{yang2019xlnet}, 
% %text classification~\cite{devlin2019bert, yang2019xlnet}, 
% sentiment analysis~\cite{yang2019xlnet} and more.
% %and general language understanding~\cite{devlin2019bert}. 
%
% Attention based models have revolutionized machine translation and natural language processing ~\cite{bahdanau2015neural, vaswani2017attention, wu2018pay, devlin2019bert, dai2019transformer, yang2019xlnet}.} \ran{from: Exploring self attention ~\cite{zhao2020exploring}}
%
% In particular, self-attention~\cite{cheng2016long, } calculates the response at a position in a sequence by attending to all positions within the same sequence.
% Vaswani et al. ~\cite{vaswani2017attention} demonstrated that machine translation models could achieve state-of-the-art results by solely using a self-attention module.
%  \ran{from: ~\cite{zhang2019self}}
%
% 2D
The success in NLP has inspired applications of attention to image analysis tasks.
These include, among others, recognition~\cite{dosovitskiy2020image, hu2019local, zhao2020exploring}, 
%~\cite{bello2019attention, dosovitskiy2020image, hu2019local, Ramachandran2019standalone, wang2017residual, zhao2020exploring, zhao2018psanet}, 
image synthesis~\cite{zhang2019self}, 
%~\cite{brock2018large, zhang2019self}, 
and image captioning~\cite{xu2015show, yang2016stacked}.
%, and video prediction~\cite{jia2016dynamic, wang2018non}. 
% \ran{from: ~\cite{zhao2020exploring}}.
%
% Attention has been vastly utilized in images as well, for a variety of tasks, including among others image captioning~\cite{xu2015show}, question answering~\cite{yang2016stacked}, image generation~\cite{zhang2019self}, classification~\cite{wang2017residual, zhao2020exploring}, object detection, semantic segmentation~\cite{chen2016attention}, and scene segmentation~\cite{fu2019dual}.

%
% ==== Multi-views ====
In 3D, many of the works utilize attention within the multi-view representation, either aggregating features by attention in consecutive views~~\cite{han20193d2seqviews,he2019view,wei2020view} or, in addition, selecting the next views according to attention~~\cite{chen2018veram,han2018seqviews2seqlabels}.
% In particular, for multi-views representation, 3D2SeqViews~\cite{han20193d2seqviews} and VNN~\cite{he2019view} apply a view-wise convolution on consecutive view sub-sequences in a circular trajectory and then aggregate features by attention.
%Sequential views are selected and aggregated by RNN with attention in VERAM ~\cite{chen2018veram} and Point2Sequence~\cite{han2018seqviews2seqlabels}. 
%\ran{View-GCN~\cite{wei2020view} utilizes dynamic graph convolution operations and non-local message passing to learn discriminative shape descriptor}
% for 3D shape classification & retrieval.
% \ran{from:~\cite{wei2020view}}
% In particular, for multi-view 2D projections, attention is used to learn the importance of the different views~\cite{chen2018veram,han2018seqviews2seqlabels}.
%
% ===== Point clouds =====
Recently, attention has been used also for point clouds, 
%for a variety of applications, including retrieval, object detection, segmentation and classification.
attempting to capture the local context of a point~\cite{hu2020randla,zhang2019pcan} or the global context~\cite{paigwar2019attentional,sun2020acne}.
Others have learned contextual relation between point patches~\cite{xie2020mlcvnet}.
Point transformers have also been proposed~\cite{xie2018attentional,yang2019modeling}.
%\ran{from:~\cite{sun2020acne}}.
%\ran{In RandLA net~\cite{hu2020randla}, scene segmentation is performed using local spatial encoding to explicitly extract local geometric patterns which are aggregated utilizing attentive pooling.} 
%\ran{ACNe~\cite{sun2020acne} utilize local and global attention to find the essential data points for camera pose estimation \& classification under noise and outliers.} 
%\ran{
%For object detection, ~\cite{paigwar2019attentional} propose attentional PointNet for searching regions of interest in large-scale point clouds;
%~\cite{xie2020mlcvnet} utilize self-attention to learn multi-level contextual relation between points patches, objects and a whole scene of point cloud.} 
% \ran{
%Point transformer~\cite{zhao2020point}, Point Attention Transformer~\cite{yang2019modeling} and ~\cite{xie2018attentional} utilize self-attention~\cite{vaswani2017attention} for point clouds, which is invariant to the permutation of the point set, making it suitable for point set processing tasks such as classification \& segmentation.}

% ===== Meshes =====
The work on mesh attention, however, is sparser.
It was used for reconstructing 3D human pose~\cite{lin2020end}, for mesh deformation~\cite{yang2020multiscale}, or for classification and segmentation~\cite{Milano20NeurIPS-PDMeshNet}.
In PD-Meshnet~\cite{Milano20NeurIPS-PDMeshNet}, the primal-dual graph framework is extended to 3D meshes, utilizing graph attention network to capture global context. 
%~\cite{velickovic2018graph} 
The non-local nature of transformers is exploited in~\cite{lin2020end}. 
%\ran{from: arxiv survey} %https://arxiv.org/pdf/2012.12556.pdf
% cite{yang2020multiscale} new arxiv paper with 1 citation 
Attention masks are extracted in~\cite{yang2020multiscale} in order to attend different shape parts at lower scale, enabling fine part-deformation for local attentive regions.

Our proposed attention is inherently different.
It operates on multiple random walks on a mesh, using attention to learn to produce a single cross-walk attentive features.

%%%%%%%%%%%%%%%%%%%%%%%%%%%%
%%%% section: Model
%%%%%%%%%%%%%%%%%%%%%%%%%%%%
\section{Model}
\label{sec:model}

% goal
We wish to learn how to separate the wheat from the chaff, focusing on the relevant features of the mesh and ignoring the irrelevant ones.
For instance, to distinguish a stool from a chair in Figure~\ref{fig:teaser}, it is useless to explore their seats; rather it is useful to identify the backrest (or the lack of it).
Generally, in Figure~\ref{fig:teaser} it is better to focus on the features along the cyan walk than on features along the magenta walk.
%

% Architecture figure
\begin{figure*}[tb]
\begin{center}
\includegraphics[width=0.9\linewidth]{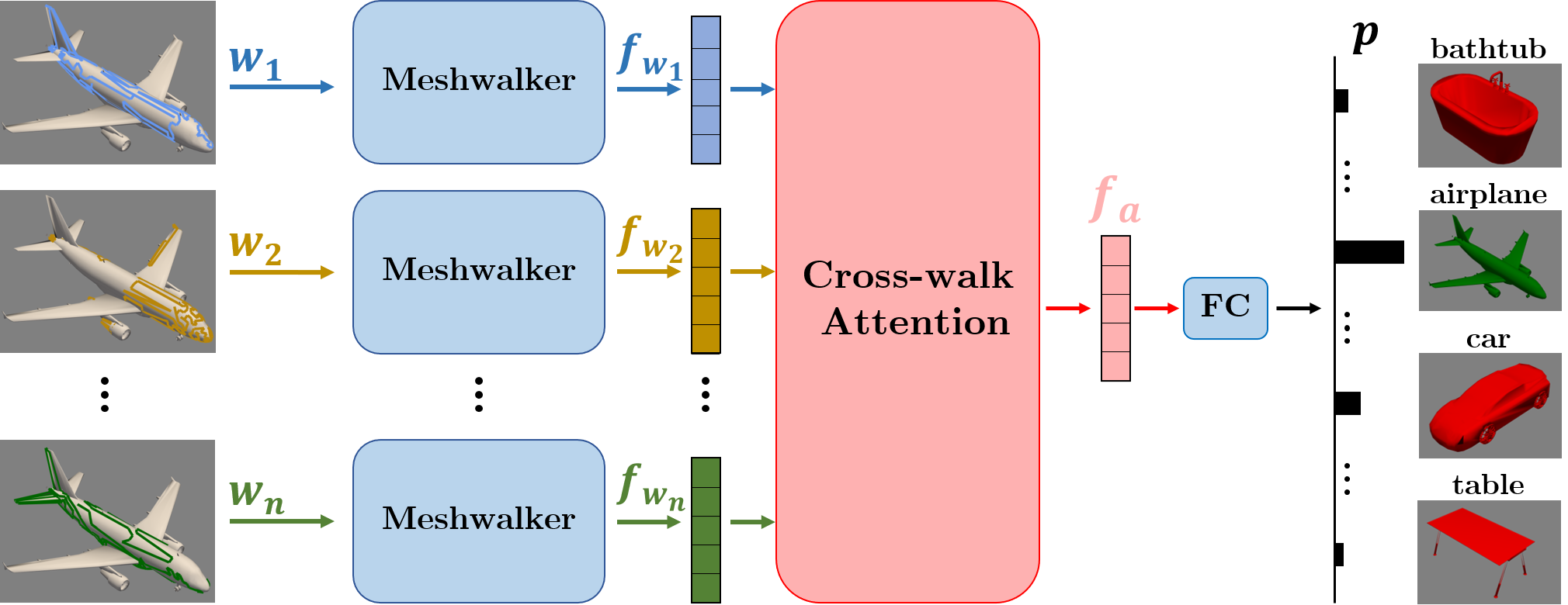}
\end{center}
   \caption{{\bf Architecture.} 
   Each walk, $w_i$, is processed independently by {\em Meshwalker}~\cite{lahav2020meshwalker},  excluding its last classification layer, resulting in a feature vector, $f_{w_i}$, for each walk.
   These $n$ feature vectors are the input to our novel attention module, which produces a single mesh feature vector, $f_a$, which emphasizes the most attentive properties of the mesh.
   The last fully-connected layer transforms $f_a$ into a probability vector, which is used for shape analysis applications (e.g. $p$ is a prediction vector for classification).
   Figures~\ref{fig:meshwalker} and \ref{fig:attention} illustrate the architectures of MeshWalker and the attention module. 
   }
\label{fig:full_architecture}
\end{figure*}

% Key idea
We propose a novel model that makes use of the fact that we have multiple pieces of information.
In particular, we benefit from having multiple random walks, which explore the mesh in diverse manners.
Briefly, a random walk is defined as a series of vertices, whose first vertex is selected randomly, and then the next vertices are added iteratively, where each vertex is chosen randomly from the vertices adjacent (along a joint edge) to the current one. 
Each vertex of a walk is represented as the 3D translation from the previous vertex.
%The information accumulated along the walk is distillates into a single descriptor vector using a {\em Recurrent Neural Network (RNN)} architecture that "remembers" the walk's path.
%
Thus, each walk wanders around the mesh, going through meaningful, as well as non-meaningful parts of the mesh. 
% non-distinctive areas.
In MeshWalker~\cite{lahav2020meshwalker}, for each random walk a different feature vector is learned, describing the mesh from the specific walk's perspective.
% Some features extracted by the walk are significant, while others are less significant.

%Given various descriptions of the same mesh enables us to take into account the most informative (parts of) the descriptors.
%
Our key idea is that, given multiple feature vectors, learned from their respective walks, we will learn to focus on the most informative entries of these descriptors.
This is done by a novel {\em Many-to-One (MtO)} attention module that, given information from multiple sources, generates a single feature vector.
This descriptor highlights the informative features and neglects the others.

% analogy
In analogy to multiple sentences that describe the same event, we are given multiple  walks that describe the same mesh.
%Though each sentence is ordered, similar words of the different sentences need not necessarily appear in the same location.
Each sentence is processed separately and a feature vector is generated for it, describing the event as expressed by this sentence.
Our goal is to learn a single feature vector that describes the event as a whole, by utilizing the collection of feature vectors generated for the sentences.
This is different from from common attention units~\cite{vaswani2017attention}, where attention is used to improve the individual descriptors. 
We note that this analogy is not complete, as randomness is unique to our case.

% attention module - idea
Specifically, our attention module gets as input $n$ feature vectors for $n$ walks.
We may think of every vector's entry as describing a certain property of the walk.
Thus,  in order to compute entry $i$ in the resulting mesh feature vector, all $i$-s entries of the walk feature vectors are used, independently of the other vectors' entries.
This is done by learning a weight for each entry of each feature vector.
For learning these weights, however, all entries from all feature vectors are used.
%
% Ran - Also its beneficial to "stay the same" - multiwalk and single walk share the feature space - classifier for single walk can work the same as on our output.
Finally, the weight vectors, jointly with the walk feature vectors, are used to generate the final result.
In the following we will elaborate on the details of this idea.
However, we note already that our scheme may be beneficial in other scenarios where the same object/scene/event can be described in diverse, though tightly related, manners.

% Architecture - outline
Figure~\ref{fig:full_architecture} illustrates the architecture of our proposed model. 
It consists of $n$ instances of MeshWalker, excluding the last classification layer. Each instance processes a single walk, $w_i$, independently, generating a feature vector for this walk $f_{w_i}$, $1 \leq i \leq n$.
These feature vectors constitute the input to our cross-walk attention module, which generates a single feature vector that describes the mesh, $f_a$. 
The final prediction vector, $p$, is generated by a fully-connected layer, which gets $f_a$ as input . 
Hereafter, we briefly describe MeshWalker and then elaborate on the attention module, which is the key of the framework.

\begin{figure}[tb]
\begin{center}
\includegraphics[width=1\linewidth]{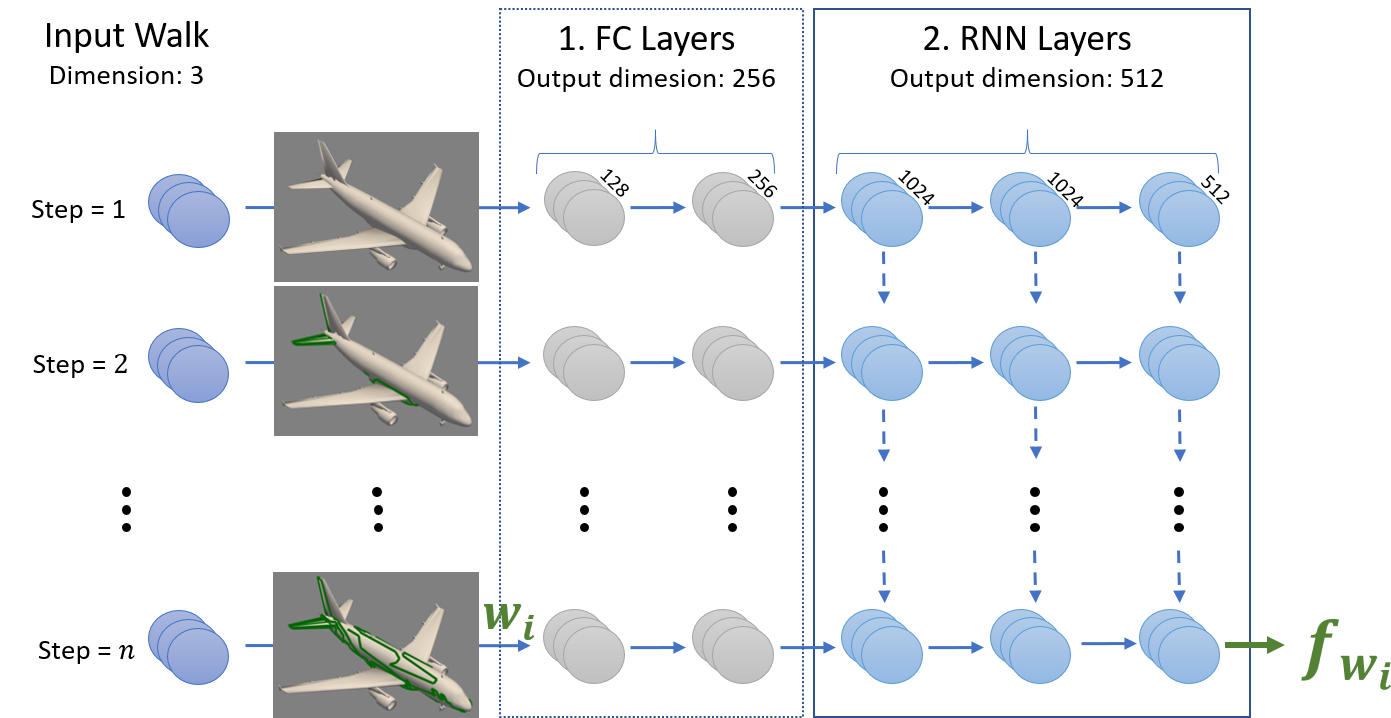}
\end{center}
   \caption{\textbf{Meshwalker~\cite{lahav2020meshwalker}.}
   This network gets as input a random walk (a sequence of vertices) along the mesh, $w_i$.
   Each vertex is first embedded into a higher dimension feature vector by  two fully-connected layers. 
   Then, subsequent three RNN (GRU) layers process the sequence of feature vectors into a single walk feature vector, $f_{w_i}$, which describes the properties of the walk.
   }
\label{fig:meshwalker}
\end{figure}

% Meshwalker
\vspace{0.05in}
\noindent
{\bf MeshWalker.}
As illustrated in Figure~\ref{fig:meshwalker}, MeshWalker~\cite{lahav2020meshwalker} gets as input the 3D coordinates of the vertices along the walk. 
It first learns to map each vertex to a new feature space in high dimension by fully-connected layers. 
%This is performed by two fully-connected (FC) layers followed by an \textit{instance normalization} layer and \textit{ReLu} activation.
Then, a \textit{Recurrent Neural Network (RNN)}, having a hidden state vector ("memory") that contains the information gathered along the walk, is applied.
It learns to accumulate the important information along the walk and forget the non-important information.
% This network has a hidden state vector, "memory", which describes the walk up to the current vertex, and outputs a state vector that contains the information gathered along the walk.
The RNN is implemented using three \textit{Gated Recurrent Units (GRU)}~\cite{cho2014learning}.

% Meshwalker figure %%%%%%%%%%%%%%%%%%%

% =============== Attention figure =============== %
\begin{figure*}[tb]
\begin{center}
\includegraphics[width=1\linewidth]{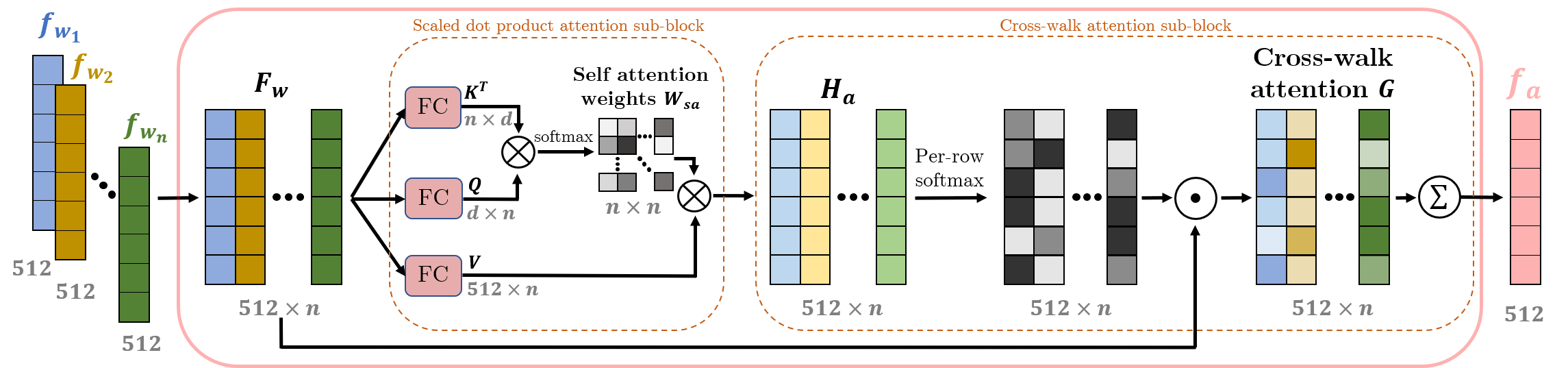}
\end{center}
   \caption{
   \textbf{Cross-walk attention.}
   Given $n$ feature vectors $f_{w_i}$, each representing a walk $w_i$,  we compute a cross-walk attention vector $f_a$.
%   This is done by learning the weight for each entry of the feature vectors.
   At first, the {\em scaled dot product attention} of~\cite{vaswani2017attention} is applied to the input vectors.
  It starts by utilizing $3$ parallel fully-connected layers per walk:
  the first two learn the attention between each walk vector to the other walk vectors, 
%  using matrix multiplication (denoted by $\bigotimes$);
  and the third transfers the input walk to be multiplied by a function of the former two.
  The output of this sub-block are attention feature vectors for the walks , denoted as $H_a$.
   The second sub-block generates a single vector that represents the mesh in a way 
   that weighs the importance of each walk to each entry.
   % each entry considers the walks relevant to it and neglect the rest.
   This is done in three steps, given $n$ walk attention feature vectors~$H_a$:
   (1)~{\em softmax}  is applied per row in order to transform it into a weight (probability) vector.
   % This vector represent the importance of each walk to the specific row feature.
   (2) Hadamard product~($\bigodot$) between the acquired weights and the input walk features scales each feature entry according to its learned importance.
   (3) The weighted walk features are summed across the walks, to produce the output feature vector $f_a$.
%  This vector represents the mesh in a way such that each entry considers the walks relevant to it and neglect the rest.
   }
\label{fig:attention}
\label{fig:twocol}
\end{figure*}

% 2nd block
\vspace{0.05in}
\noindent
{\bf Cross-walk attention.}
Our proposed cross-walk attention block is illustrated in Figure~\ref{fig:attention}.
Given $n$ feature vectors of $n$ walks, the attention module learns how to aggregate the important information from all the walks, while discarding the irrelevant information.
This block consists of two sub-blocks.
In the first, the multiple walks are utilized in order to produce a better description for each walk separately, generating an individual attentive vector per walk.
Then, these attentive vectors are put together in a way that provides a single attentive vector for the whole mesh.

% first sub-block
In particular, for the first sub-block,  we are given $n$ walk feature vectors $\{f_{w_i}\}_{i=1}^n$, stacked to form a matrix~$F_w$. 
The goal is to generate a new matrix of walk features, $H_a$, in which each walk is enriched by information from other walks. 
Given $F_w$, we learn its self-attention map, using the scaled dot-product attention block of~\cite{vaswani2017attention}, which
transforms each walk features into a self-attention vector.
%, giving a different score for each of its entries. 
%
Briefly, $F_w$ is transformed into three feature sub-spaces $Q,K,V$.
Intuitively, the columns of $Q$ and $K$ are walks descriptors, learned to represent relevancy to one another.
While~$Q$ and~$K$ play a similar role, their different walk descriptors enable the walks to influence one another in a non-symmetric manner.
%hold similar interpretation, their differences allow non-symmetric relations between walks.
$V$'s columns represent intra-walk attention per entry.
$Q,K,V$ are computed as follows:
\begin{equation}
 Q=W^{(q)}F_w, \hspace{0.1in}
K=W^{(k)}F_w, \hspace{0.1in}
V=W^{(v)}F_w ,
\end{equation}
% \begin{eqnarray}
%  \nonumber    Q&=&W^{(q)}F_w \\ 
%              K&=&W^{(k)}F_w \\ 
% \nonumber    V&=&W^{(v)}F_w  
% \end{eqnarray}
where $W^{(q)},W^{(k)}, W^{(v)} \in \mathbb{R}^{d \times d}$ are learned weight matrices that are used to project the  walk feature vectors into different sub-spaces of dimension $d$ (in our implementation $d=512$, the same dimension as that of the walk feature vectors).
 % $Q, K, V$ enable the model to learn the relations between walks, termed as the self-attention weights. 
The self-attention weights are computed as
\begin{equation}
W_{sa}= softmax(\frac{QK^T}{\sqrt{d}}).
   \label{eq:Wsa}
\end{equation}
Finally,  the self-attention feature vectors are aggregated in matrix $H_a$.
Each column of $H_a$ utilizes knowledge from all the walks and 
% why H_a is not sufficient
will be used next to weigh the importance of the different features of each walk, in order to represent the mesh as a whole.
$H_a$ is defined as
\begin{equation}
 H_a=W_{sa}V.
   \label{eq:Ha}
\end{equation}
%function similar to \ayellet{Query} and \ayellet{Keys} in retrieval frameworks; \ayellet{their goal is to find pair-wise similarities (attention) between two sets.}
% \ayellet{ In our model, the two sets are the same set of walk features learned "attributes".
%  By the same analogy, $V$ are the values to retrieve, which are learned per-walk features the attention will linearly combine. }
%
% The output of this block, $H_a$, utilizes these relations.
%  Note that $W^{(v)}$ must be set to the same dimension of $F_w$. 

% second sub-block
The second sub-block (the cross-walk attention sub-block in Figure~\ref{fig:attention}) aims at aggregating the walk features in~$F_w$ into a single mesh feature vector $f_a$.
This is done by utilizing $H_a$, in which each entry indicates the importance of that entry.
$H_a$ is first given a probabilistic interpretation, by applying softmax  per row $i$.
%for each row (feature), resulting in cross-walk attention weights.
These probabilities are then multiplied, element-wise, by the walk feature vectors~$F_w$, to create cross-walk attention features.
That is to say, every row is weighted (cross-walk) according to the learned probabilities of each walk.
Finally, the columns $j$ of the cross-walk attention matrix are summed, creating the sought-after mesh feature vector, $f_a$.
This procedure is expressed as
% \begin{equation}
%     (f_a)_i = \sum_{j=1}^n (softmax(H_a)_i\odot F_w)_\ayellet{{i,j}}.
%     \label{eq:Fa}
% \end{equation}
\begin{eqnarray}
      &&G = F_w \odot softmax(H_a)\\
     &&(f_a)_i = \sum_{j=1}^n G_{i,j} \nonumber
   \label{eq:Fa}
\end{eqnarray}

Note that $f_a$ resides in the original walk feature sub-space, since each of its entries is a linear combination of the corresponding entries of $F_w$.
This enables the classification layer in Figure~\ref{fig:full_architecture} to process both single-walk features and cross-walk attentive features seamlessly.
This is essential to our two-step training method, which is described next.

% ===============
%Other possible solutions to aggregate information from multiple walks include average, max/min pooling, and dot product attention. 
%Average pooling accumulate the information across walks with equal importance. max/min pooling focus on a single walk, discarding both important and irrelevant information which other walks may contain. 
%Dot product attention give equal per walk weighting for all channels of the feature vectors, thus lack the fine per-entry weighting we suggest. 

%===============
% We note that our method achieve per walk per entry weighting of the input features without adding parameters over the dot product attention.
% \ayellet{Explain this sentence}
% \ran{Since we interpret the scaled dot product attention output as cross-walk probabilities by a simple softmax operation, it allow us to achieve per walk per entry weighting of the input features, allowing us to multiply-add aggregate multiple walks into a single mesh feature vector.}

%%%%%% Training
\vspace{0.05in}
\noindent
{\bf Training.}
Our model is trained in two phases. 
At first, the network is trained without our cross-walk attention, letting the model learn to extract meaningful features for each walk independently, attempting to correctly classify the shape by every single walk.
This assists in extracting the best features per walk.
%
%, instead of always  attending the most informative walk (the "easy" way) which might not be sampled or even not exist in some of the test models. We show this scheme leads to improvement in results.
In the second phase, we freeze the MeshWalker block (the cyan rectangles in Figure~\ref{fig:full_architecture}) and train the attention block, so as to account for the most relevant features across all walks.
The two-phase training prevents the network from focusing only on the most attentive walks, which would result in avoiding to process mesh regions that are less attentive, but may be important for fine results. 
% \ran{Refer to Ablation Study for further training comparison.}
% They note that it leads to better performances (su2015multi). the others quote them when writing they train in two-steps.
In Section~\ref{sec:ablation} we compare this strategy to 1-phase training.

Recall that both the individual walk features $f_{w_i}$ and the mesh feature vector
 $f_a$ reside in the same sub-space, thus both phases can be trained by the same loss.
Given a prediction vector (either per-walk in Phase 1 or $p$ of Figure~\ref{fig:full_architecture} in Phase 2) and the corresponding class label $l$, 
for mesh classification, training is performed by minimizing the Softmax cross entropy loss
\begin{equation}
    L(p,l)=-\log{\frac{e^{p_l}}{\sum_{j=1}^{C} e^{p_j}}}.
    \label{eq:loss}
\end{equation}
%where, $p$ is the prediction vector (Figure~\ref{fig:full_architecture}) and $l$ is the true label $l\in[1, 2,.., C]$, where $C$ is the number of classes. 

For retrieval, we train our model using a combination of the softmax cross entropy loss and the triplet-center loss (\textit{TCL})~\cite{he2018triplet}. 
Intuitively, {\em TCL} attempts to push each walk prediction vector closer to its corresponding class center and away from centers of other classes.
%pulling features from the same category closer together and pushing features from different categories apart.}
% It does not effect the classification much but improve retrieval significantly.
%
%
Specifically, given $p$ and $l$ as above, the TCL is defined as 
\begin{equation}
    TCL(p, l) = \max{(D(p, c_l) + m - \min_{k\ne l}D(p, c_k), 0)}.
\label{eq:TCL}
\end{equation}
Here, each class is represented by a learned parametric center $c_{k}$ (of the same dimension as $p$). 
The margin $m$\ is a hyper parameter that prevents pushing the vector too far ($m=1$ in our experiments).
$D(\cdot)$ is the squared Euclidean distance.
The combination of the losses is defined as
\begin{equation}
    \mathcal{L}(p, l) = \lambda_1TCL(p, l)+ \lambda_2L(p,l).
\label{eq:TCL}
\end{equation}
In all our experiments $\lambda_1=1$ and $\lambda_2=0.01$.
We note that though this loss encourages the network to learn more discriminative mesh features and indeed improves retrieval results, empirically it does not improve classification results.

% In the first phase of retrieval training, we use the average of the prediction vectors of the individual walks $\frac{1}{M}\sum_{i=1}^{M}p_i$.
% In the second training phase, we apply the exact same procedure, this time using the mesh prediction vector $p$, calculated from $f_a$ (Figure~\ref{fig:full_architecture}). 

In practice, training is performed in batches of  $M=64$ walks, where a batch contains walks from several meshes.
In the first phase each walk belongs to a different mesh, whereas in the second phase we use $8$ walks per mesh, for  $8$ meshes.
All the meshes in our experiments are first simplified into $1K$, $2K$ and $4K$ faces, both to reduce the network capacity required for training and as a form of data augmentation.
At inference, we average the scores of the predictions at the different scales.
Meshes with less faces than the above scales are used without simplification.
The models are normalized into a unit cube.
For our first training phase, we  use 
%the same optimization parameters as~\cite{lahav2020meshwalker} 
Adam optimizer with cyclic learning rate of $5\cdot10^{-4}$ to $10^{-6}$ with $20K$ iterations per cycle, for a total of $200K$ iterations.
%for Modelnet, ShapeNet-Core55 and 3D-FUTURE, and $80K$ for SHREC11.
In the second phase, we reduce the learning rate by half, which is more stable for fine-tuning the cross-walk attention block, training for additional $100K$ iterations.
% for Modelnet, ShapeNet-Core55 and 3D-FUTURE, and for $40K$ more for SHREC11.

%%%%%%%%%%%%%%%%%%%%%%%%%%%%
%%%% section: Applications
%%%%%%%%%%%%%%%%%%%%%%%%%%%%
\section{Applications}
\label{sec:applications}

% applications 
The performance of our model is evaluated in 3D shape classification and retrieval, on a variety of datasets.
For each dataset, we compare our results to the reported results; hence, each table shows results of different algorithms.

%%%%%%%%%%%%%%%%%%%%%%%%%%%%
% subsection: Mesh classification
%%%%%%%%%%%%%%%%%%%%%%%%%%%%
\subsection{Mesh classification}

%\ran{Itzik - run on modelnet10 to see if we could benefit from pointnet++ alignment in entrance of network}

% problem definition
Given a mesh, the goal is to classify it into one of pre-defined classes. 
For each mesh, we apply our model, as described in Section~\ref{sec:model}, where
the last fully-connected layer outputs a classification prediction vector $p$.
%
% datasets
The three datasets utilized differ in the number of classes and the number of objects per class.
We report both on \textit{instance accuracy} and on \textit{class accuracy}.
{\em Instance accuracy}  is defined as the percentage of the correctly-classified objects.
{\em Class accuracy} is defined as the mean of class instance accuracy; thus it considers all classes equally, ignoring their size.
The two metrics are the same for SHREC11, which is class-balanced, and differ for the imbalanced datasets, ModelNet40 \& 3D-FUTURE.
%
% \begin{equation*}
%     Instance = \sum_{i=1}^{N}\frac{\mathbbm{1}_{x_i=y_i}}{N}
% \end{equation*}
% Where $\mathbbm{1}$ is the indicator function and $N$ is the number of samples $\{x_i, y_i\}$ in the test set.
% \begin{equation*}
%     Class = \frac{1}{K}\sum_{k=1}^{K}\frac{\sum_{j=1}^{n_k}\mathbbm{1}_{x_{j,k}=y_k}}{n_k}
% \end{equation*}
% Where $K$ is the number of classes, $n_k$ is the number of samples $\{x_{j,k}, y_k\}$ of class $k$ in the test set.

% Results per dataset

% ================= 3D-FUTURE ============= %
\vspace{0.05in}
{\bf 3D-FUTURE~\cite{fu20203d}}. 
This new dataset contains $9,992$ industrial CAD models of furniture~\cite{fu20203d}.
It consists of $7$ super-categories, having $1$-$12$ sub-categories each, for a  total of $34$ categories.
The train/test split is $6,699$/$3,293$.
This dataset is challenging both due to the objects it contains and due to its hierarchical structure, as objects in related sub-categories may resemble each other, requiring fine-grain classification.

For this dataset, we trained our model with the {\em class-balanced loss} of~\cite{cui2019class}, which was found empirically to outperform cross-entropy.
This loss handles well heavily-unbalanced datasets (the number of training shapes per category ranges between $8$ 
%(\emph{classic chinese chair}) 
to $633$).
%(\emph{three-seat sofa})
It is given by: 
\begin{equation}
    CB_{softmax}(p,l)=-\frac{1-\beta}{1-\beta^{n_l}}\log{\frac{e^{p_l}}{\sum_{j=1}^C{e^{p_j}}}}.
\label{eq:classbalanced}    
\end{equation}
For each mesh prediction vector $p$ and label $l$, we weigh the cross entropy loss according to the number of training objects with the same label $n_l$, where $\beta$ is a hyper-parameter in the range $[0, 1]$, set empirically to $0.9$.

Table~\ref{tab:future} shows that our method outperforms previous methods, both point-based or multi-view.
Following~\cite{fu20203d}, we omit categories with less than $10$ training samples from the train/test sets, thus left with $32$ categories.

% ================= 3D-Future classification table ============= %
\begin{table}[t]
\begin{center}
\begin{tabular}{|l|c|c|c|c|}
\hline
Method & Input & Class & Instance  \\
\hline\hline
AttWalk (Ours)                                  & mesh & ${\bf 72.1\%}$ & ${\bf 73.7\%}$ \\
% MeshWalker~\cite{lahav2020meshwalker}      & mesh & 68.2\% & 70.5\% \\
% MeshNet~\cite{feng2019meshnet} & mesh & 64.0\% & 63.3\% \\
\hline\hline
% KPConv~\cite{thomas2019kpconv} & point cloud &  69.3\% & 69.9\% \\
PointNet++~\cite{qi2017pointnet++}   & point cloud &  $69.9\%$ & -  \\
\hline\hline
MVCNN~\cite{su2015multi}       & multi-views & $69.2\%$ & - \\
\hline
\end{tabular}
\end{center}
\caption{{\bf 3D-FUTURE classification (class/instance accuracy).} 
Our results outperform those reported in~\cite{fu20203d}.
%, whereas KPConv \& MeshNet are trained using the respected provided codes.
}
\label{tab:future}
\end{table}

% ================= SHREC 11 ============= %
\vspace{0.05in}
{\bf SHREC11~\cite{lian2011shape}.}
This dataset consists of $30$ classes, each contains $20$ meshes. 
Typical classes are camels, cats, glasses, centaurs, hands etc. 
Following the setup of~\cite{ezuz2017gwcnn}, the objects in each class are split into $16$ (/$10$) training examples and $4$ (/$10$) testing examples.

Table~\ref{tab:shrec11} compares the performance of state-of-the-art algorithms on this dataset.
Each result is the average of $3$ random splits into train/test sets. 
Our method outperforms SOTA methods.
In fact, for the $16$/$4$ split it achieves a perfect score and for the $10$/$10$ split an almost-perfect score.
%\ayellet{no results for multi-view on this dataset?}

% ================= SHREC 11 classification table ============= %
\begin{table}[tb]
\begin{center}
\begin{tabular}{|l|c|c|c|}
\hline
Method & Input & Split-16 & Split-10 \\
\hline\hline
AttWalk (Ours)            & Mesh & $\textbf{100\%}$     & $\bf{99.7\%}$ \\
PD-MeshNet~\cite{Milano20NeurIPS-PDMeshNet}  & Mesh & $99.7\%$  & $99.1\%$ \\
MeshWalker \cite{lahav2020meshwalker} & Mesh & $98.6\%$  & $97.1\%$ \\
HSN~\cite{wiersma2020cnns}            & Mesh &  -      & $96.1\%$ \\
MeshCNN \cite{hanocka2019meshcnn}     & Mesh & $98.6\%$  & $91.0\%$ \\
GWCNN~\cite{ezuz2017gwcnn}            & Mesh & $96.6\%$  & $90.3\%$ \\
SG~\cite{bronstein2011shape}          & Mesh & $70.8\%$  & $62.6\%$ \\
\hline
\end{tabular}
\end{center}
\caption{{\bf SHREC11 classification.}
Split-16(/10) indicates that $16$(/$10$) objects were used for training out of 20 in each class. 
Our method achieves perfect results for the $16/4$ split and almost perfect results for the $10/10$ split.
}
\label{tab:shrec11}
\end{table}

\vspace{0.05in}
{\bf ModelNet40~\cite{wu20153d}.} 
This dataset contains $12,311$ CAD models from $40$ categories, out of which $9,843$ models are used for training and $2,468$  for testing. 
This dataset contains many non-watertight and multiple-component objects, which might be difficult for some mesh-based methods.

Table~\ref{tab:modelnet_classification} shows that our method outperforms other mesh-based methods.
We note that while in~\cite{lahav2020meshwalker}, $64$ walks are used, in our framework $8$ walks per shape suffice.
This is thanks to our cross-walk attention that focuses on the relevant information from each walk and neglects the non-informative features.
Our results are still not as good as those of multi-view methods.
This is due to relying, in addition to the mesh dataset, also on networks that are pre-trained on a large number of images~\cite{su2018deeper}.

% ================= ModelNet40 classification table ============= %
\begin{table}[t]
\begin{center}
\begin{tabular}{|l|c|c|c|}
\hline
Method & Input & Class & Instance \\
\hline\hline
AttWalk (Ours)  & mesh & $89.9\%$ &  $\textbf{92.5\%}$ \\
% MeshWalker (8 walks) \cite{lahav2020meshwalker} & mesh & 89.3\% &  92.0\%  \\
MeshWalker \cite{lahav2020meshwalker} & mesh & $89.9\%$ &  $92.3\%$ \\
MeshNet \cite{feng2019meshnet} & mesh & - & $91.9\%$  \\
\hline\hline
RS-CNN \cite{liu2019relation} & point cloud  & -    & $\textbf{93.6\%}$ \\
KPConv \cite{thomas2019kpconv} & point cloud & -    & $92.9\%$ \\
PointNet \cite{qi2017pointnet} & point cloud & $86.2\%$ & $89.2\%$ \\
\hline\hline
Subvolume~\cite{qi2016volumetric} & volume & - & $\textbf{89.2\%}$ \\
3DShapeNets~\cite{wu20153d} & volume & $77.3\%$ & $84.7\%$  \\
\hline\hline 
View-GCN~\cite{wei2020view} & multi-views & $\textbf{96.5\%}$ & $\textbf{97.6\%}$ \\
MVCNN-New~\cite{su2018deeper} & multi-views & $92.4\%$ & $95.0\%$ \\
% 3D2SeqViews \cite{han20193d2seqviews}         & multi-views & 93.4\% \\
Rotationnet~\cite{kanezaki2018rotationnet} & multi-views & $92.4\%$ & $94.8\%$ \\
\hline
\end{tabular}
\end{center}
\caption{{\bf ModelNet40 classification.} 
Our method achieves state-of-the-art results compared to other mesh-based methods; 
comparable with MeshWalker, it does so with $\frac{1}{8}$ of the walks ($64$ vs. $8$).
Multi-view methods, which are pre-trained on numerous images, are better for this dataset.
}
\label{tab:modelnet_classification}
\end{table}

%%%%%%%%%%%%%%%%%%%%%%%%%%%%%
%%%%% subsection: Retrieval
%%%%%%%%%%%%%%%%%%%%%%%%%%%%%
\subsection{Retrieval}

% Problem definition
Given a query object, the goal is to retrieve objects from a given dataset, ordered by their relevancy to the query.
Relevancy is determined for each returned object according to the query's category and sub-category (if exists).
%
%
% Datasets
We evaluate our method on two large-scale retrieval datasets: ModelNet40 and ShapeNet-Core55.
%
% Evaluation
The most common evaluation measure is the {\em mean average precision (mAP)}, which is used almost solely for ModelNet40.
% In fact, for ModelNet40, this is the only measure used in the literature 
% \ran{some use AUC- area under the curve for the other ModelNet40 split (None of who we compare with)}.
% For each dataset, we use the same evaluation metrics used in the literature.
%Specifically, for ModelNet40 we use {\em mean average precision (mAP)}.
For ShapeNet-Core55, other measures are utilized as well, most notably the {\em Normalized Discounted Cumulative Gain (NDCG)}.
%For ShapeNet Core55, we use several evaluation metrics: 
%{\em Precision at N (P@N)}, {\em Recall at N (R@N)}, {\em F1 at N ($F_1@N$)}, {\em mean average precision} and {\em Normalized Discounted Cumulative Gain (NDCG)}.
%Hereafter we briefly describe each of these evaluation measures.
%
Specifically, for a returned list with $N$ objects, we consider those that belong to the query's  category as positives and the others as negatives.
% P@N is the percentage of positive retrieved objects out of $N$.
mAP is the mean of the precision scores at every positive retrieved object position in the list.
% R@N is the percentage of  the positive retrieved objects relative to the number of objects in the category.
% $F_1@N$ is the harmonic average of precision and recall at position $N$.
For NDCG, the relevancy of each returned object is graded between $0$ to $3$, considering both category and sub-category~\cite{jarvelin2002cumulated}.
%Given retrieved list, NDCG produce a score relative to the best possible retrieved list (a list sorted from most to least relevant objects).
In~\cite{savva2017large}, both macro and micro average results are evaluated.
The macro-average gives equal weights to the scores of all the queries;  
the micro-average first averages the scores of each category and then averages the scores of the categories, giving every category an equal weight, regardless of its size.

% Results per dataset

% ModelNet40
\vspace{0.05in}
{\bf ModelNet40~\cite{wu20153d}.}
We use the most common $9,843$/$2,468$ train/test split (a few papers use other splits).
%and $80$/$20$ split per category.
Table~\ref{tab:modelnet_retrieval} shows that our method achieves SOTA results.
%, compared to other mesh-based methods, but image-based methods are still the best for this dataset.

% =================== ModelNet40 Full retrieval table ============== %
\begin{table}[t]
\begin{center}
\begin{tabular}{|l|c|c|}
\hline
Method & Input  & mAP \\
\hline\hline
AttWalk                 & Mesh &  ${\bf 91.2}$  \\
% AttWalk ($f_a$)                     & Mesh &  {\bf 89.9}  \\
% MeshWalker \cite{lahav2020meshwalker}           & Mesh & \ran{??} \\
MeshNet \cite{feng2019meshnet}                  & Mesh & $81.9$ \\
GWCNN~\cite{ezuz2017gwcnn}                      & Mesh & $59.0$ \\
\hline
\hline
DensePoint  \cite{liu2019densepoint}            & Point Cloud & ${\bf 88.5}$ \\
\hline
\hline
MVCNN~\cite{su2015multi}                        & multi-views & $79.5$ \\
SeqViews~\cite{han2018seqviews2seqlabels}       & multi-views & ${\bf 89.1}$ \\
\hline
\end{tabular}
\end{center}
\caption{{\bf ModelNet40 retrieval.} 
Our method outperforms other methods applied to the full dataset.
%, with $9843$/$2468$ train/test split.
}
\label{tab:modelnet_retrieval}
\end{table}

% =================== ModelNet40 80-20 retrieval table ============== %
% \begin{table}[h]
% \begin{center}
% \begin{tabular}{|l|c|c|}
% \hline
% Method & Input  & mAP \\
% \hline\hline
% AttWalk               & Mesh & {\bf 91.1}     \\
% % AttWalk ($f_a$)                & Mesh & {\bf 91.1}     \\
% % MeshWalker \cite{lahav2020meshwalker}           & Mesh & 90.7 \\
% \hline
% Point2SpatialCapsule \cite{wen2020point2spatialcapsule} & Point Cloud & {\bf 89.4} \\
% \hline
% MVCNN \cite{su2015multi}                        & multi-views & 79.5 \\
% GVCNN \cite{feng2018gvcnn}                      & multi-views & 85.7 \\
% TCL \cite{he2018triplet}                        & multi-views & 88.0 \\
% SeqViews2SeqLabels \cite{han20193d2seqviews}    & multi-views & 89.09 \\
% DRCNN \cite{sun2020drcnn}                       & multi-views & {\bf 93.9} \\
% \hline
% \end{tabular}
% \end{center}
% \caption{{\bf ModelNet40 retrieval} using~\cite{wu20153d} train/test split \ayellet{which is?}.}
% \label{tab:modelnet_80_20_retrieval}
% \end{table}

\begin{table}[t]
\begin{center}
\begin{tabular}{ |l c c| c c|}
\hline
 &  \multicolumn{2}{c|}{microAll} & \multicolumn{2}{c|}{macroAll} \\
% \cline{2-11}
\hline
Method  & mAP & NDCG & mAP & NDCG \\
\hline
AttWalk    & $\bf{81.1}$ & ${\bf86.7}$ & ${\bf 65.5}$ &  ${\bf68.2}$ \\
% AttWalk ($f_a$)          & Mesh & 79.6 & 86.4 & 61.4 & 65.5 \\
% MeshWalker \cite{lahav2020meshwalker}  & Mesh & 86.0 & 90.0 &  76.6 & 85.8 \\
DLAN~\cite{savva2017large}            & $66.3$ & $76.2$ & $47.7$ & $56.3$ \\
\hline
\hline
ViewGCN~\cite{wei2020view}   & ${\bf 78.4}$ & $85.2$ & ${\bf 60.2}$ & ${\bf 66.5}$ \\
GIFT~\cite{bai2016gift}        & $64.0$ & $76.5$ & $44.7$ & $54.8$ \\
MVCNN~\cite{su2015multi}       & $73.5$ & $81.5$ & $56.6$ & $64.0$ \\
RotationNet~\cite{kanezaki2018rotationnet} & $77.2$ & ${\bf 86.5}$ & $58.3$ & $65.6$ \\
                                           
% Following paper published with old SHREC2017 evaluation code, wrong results. no released code and no released info so cannot re-run with corrected code
% DeepGM \cite{luciano2018geodesic}   & 78.4 & 73.2 & 69.6 & 93.6 & 96.5 
%                & 85.4 & 45.9 & 52.3 & 92.2 & 95.8 \\
\hline
\end{tabular}
\end{center}
\caption{{\bf ShapeNet-Core55 retrieval~\cite{savva2017large}.} 
Our method outperforms both mesh-based (upper table) and multi-view (lower table) methods.}
\label{tab:shapenet_retrieval}
\end{table}

%ShapeNet Core55
\vspace{0.05in}
{\bf ShapeNet-Core55}. 
This dataset, which is a subset of ShapeNet, contains $51,162$ 3D objects from $55$ categories, each is subdivided into $1$ to $28$ sub-categories.
The dataset is split into $35,764$ / $5,133$ / $10,265$ training / validation / testing objects.
% For our experiments we use the normal version of this dataset, where the shapes are aligned.
The results are reported on the test set, using the evaluation code provided by~\cite{savva2017large}.
As in~\cite{savva2017large}, we retrieve up to $1000$ object whose Euclidean distance from the prediction vector of the query object is smaller than  $2m$, where $m$ is the margin hyperparameter from Equation~\ref{eq:TCL}.

Table~\ref{tab:shapenet_retrieval} shows that the performance of our method outperforms those of  state-of-the-art methods in both metrics. 
Recall that NDCG takes into account the sub-categories, thus our high scores demonstrate how well our cross-walk attention captures the objects' distinctive features.

This property can also be observed in Figure~\ref{fig:teaser}.
The bathtub and the desk are erroneously classified by~\cite{lahav2020meshwalker} as a lamp and a sink, respectively.
Our method correctly focuses on walks that depict a broader view of the objects, including their outlines, resulting in correct classification.

%%%%%%%%%%%%%%%%%%%%%%%%%%%%
%%%% section: Ablation Study
%%%%%%%%%%%%%%%%%%%%%%%%%%%%
\section{Ablation Study}
\label{sec:ablation}

\noindent
{\bf Insights on the most/least attentive walks.}
What characterizes attentive walks?
To answer this question we  analyze the results of ModelNet40 classification. % with more than $1K$ vertices.
We consider the attention rank of a walk according to its contribution to the final mesh feature vector $F_a$.
Since the contribution of a walk to each entry of $F_a$ differs, we average these contributions. 
Hereafter we discuss the affects we studied.\\
% We rank the walks from $1$ to $8$ according to their scores.
(1) All walks contribute to the final feature vector, however the contribution of the most attentive walk is $70\%$ higher than that of the least attentive walk ($17\%$ vs. $10\%$).\\
(2) The most-attentive walk is $36\%$ longer than the least attentive walk on average. %($125$ vs. $92$ on $800$-vertex long walks, normalized to the unit cube).
A possible explanation is that longer edges tend to describe the outline of the object, which is meaningful for capturing the shape of the object.\\
(3) No correlation is found between the the number of faces adjacent to the walk and the attentiveness of the walk.
However, similarly to (2), the most attentive walks "cover" more surface area (of the faces adjacent to the walk).\\
(4) The median Gaussian curvature of the most attentive walk is smaller (by $21\%$) than that of the least attentive walk.
This can be explained, similarly to (2), by the fact that attentive walks tend to describe the major parts of the object and not to focus on small (high-curvature) details.
% Coverage & 0.29611635 & 0.30161918 &0.3054398 & 0.30905113& 0.30408236 & 0.30313935& 0.29586217 & 0.2836586 \\

% =================== attended walks statistics for modelnet40 ============== %
% \begin{table}[h]
% \begin{center}
% \begin{tabular}{|c|c|c|c|c|}
% \hline
% Rank  & $1$ & $2$ & $7$ & $8$  \\
% \hline\hline
% Score &  $17.3 $ & $14.5 $  & $10.5 $ & $9.8 $  \\
% % W     &  $17.3 $ & $14.5$ & $13.1$ & $12.2$  & $11.6$ & $11.0$ & $10.5$ & $9.8$  \\
% Length &  $125$ & $119$ & $97$ & $92$  \\
% \hline
% \end{tabular}
% \end{center}
% \caption{\ran{\em{Attentive walks.} We present the average score, walk length and unique vertices visited for two most \& least attentive walks, averaged for ModelNet40 objects.
% }}
% \label{tab:stats}
% \end{table}

% 2-phase training 
\vspace{0.05in}
\noindent
{\bf Training: 2-phase vs. 1-phase.}
Recall that we train our model in two phases, first training only MeshWalker and then training only the attention module.
We compare this procedure to a single-phase (end-to-end) training, for ModelNet40 and 3D-Future.
% We use the same hyper parameters of our second-phase training for both experiments. 
% specifically using batches of size $M=64$, with $8$ meshes \& $8$ walks per mesh, similar to our second-phase training.
For both datasets, 2-phase training yields better results:
The instance accuracy is $92.5$ vs. $89.2$ for ModelNet40 and $73.7$ vs. $71.4$ for 3D-Future.
A possible explanation is that 2-phase training forces MeshWalker to learn the most meaningful features for each walk independently, whereas an end-to-end system makes it easier to focus on the "easier" (more meaningful) walks.
As we saw above, even the least-attentive walks contribute to the final results, and hence should not be neglected.

% =================== one-step vs. two-step training ============== %
% \begin{table}[h]
% \begin{center}
% \begin{tabular}{|l|c|c|}
% \hline
% Method & ModelNet40  & 3D-Future \\
% \hline\hline
% AttWalk $2$-phase       & 92.5 & 72.1 \\
% AttWalk single phase          & 89.2 & 68.4 \\
% MeshWalker \cite{lahav2020meshwalker}  & 92.3 & 69.9 \\
% \hline
% \end{tabular}
% \end{center}
% \caption{\ran{\em{two-phase training} results of our model trained end-to-end (single phase) compared to two-phase training.}}
% \label{tab:training}
% \end{table}

\vspace{0.05in}
\noindent
{\bf Number of walks.}
How many random walks suffice for optimal exploration of a mesh? 
Table~\ref{tab:walks} shows that $8$ walks already achieve the best results for SHREC11 classification.
Similar results are shown in Table~\ref{tab:modelnet_classification} for ModelNet40.

% =================== walk number SHREC11 ============== %
\begin{table}[h]
\begin{center}
\begin{tabular}{|l|c|c|c|c|c|c|}
\hline
\# Walks  &   $1$  &  $2$ &  $4$   &   $8$   &  $16$  &  $32$   \\
\hline\hline
Accuracy        & $98.1$ &  $99.1$ & $99.6$  &  $99.7$  &  $99.7$ &  $99.7$  \\
\hline
\end{tabular}
\end{center}
\caption{{\bf Number of walks.} $8$ walks suffice for best performance, in contrast to~\cite{lahav2020meshwalker}, where $32$-$64$ walks are used.}
\label{tab:walks}
\end{table}

% \ran{The effect of TCL on retrieval}
% The effect of Triplet Center Loss on retrieval results is given in Table~\ref{tab:TCL}.
% TCL significantly improves the model ability to distinguish between feature vectors ($f_a$) from different categories. 
% % =================== TCL ablation ============== %
% \begin{table}[h]
% \begin{center}
% \begin{tabular}{|l|c|c|}
% \hline
% Walks per mesh  &   ModelNet40  &    ShapeNet Core55   \\
% \hline\hline
% AttWalk w/ TCL       &   89.9    &   81.1    \\
% AttWalk wo/ TCL      &   71.4    &   79.6    \\
% \hline
% \end{tabular}
% \end{center}
% \caption{\ran{\em{Retrieval comparison.} 
% Retrieval results (mAP) of our model with and without Triplet Center Loss.}}
% \label{tab:TCL}
% \end{table}

% compare attention to max/min pooling & H_a
\noindent
{\bf Alternative walk aggregation methods.}
Our model aggregates walks using our novel attention scheme.
Table~\ref{tab:agg_compare} compares our results to alternative aggregation strategies for generating a single shape descriptor from multiple walk features.
It shows that our proposed scheme outperforms average \& max pooling on $f_{w_i}$, as well as adding average \& max pooling to the self-attention matrix $H_a$.
The latter two aggregations demonstrate that indeed self-attention by itself does not suffice for achieving SOTA results.
% the multiple walks feature vectors $f_{w_i}$, and with max-pooling the walks self attention vectors, columns of $H_a$.
% All methods utilize the same MeshWalker block trained on ModelNet40, with $8$ walks per mesh.
% For fair comparison, we trained the last method ($H_a$) exactly the same way as our second phase training.
% The results shown in Table~\ref{tab:agg_compare} support the benefits of our cross-walk attention, and demonstrate that using $H_a$ as the new walk features is not sufficient.
%TODO: Add possible reason, it should atleast remain .
% ==============  average vs. maxpool vs. attention ==== %
\begin{table}[t]
\begin{center}
\begin{tabular}{|l|c|c|}
\hline
Aggregation method & Class & Instance \\
\hline\hline
% Cross-walk Attention (our)   & $89.9$ & $92.5$    \\
Cross-walk Attention (our)   & ${\bf 72.1}$ & ${\bf 73.7}$   \\
% Average pooling             & $86.1$ & $92.0$   \\
Average pooling                  & $70.1$ & $71.0$    \\
% Max pooling                  & $88.7$ & $91.7$    \\
Max pooling                  & $58.2$ & $60.5$    \\
% $H_a$ + Average pooling      & $89.0$ & $91.7$    \\
$H_a$ + Average pooling      & $69.3$ & $71.7$    \\
% $H_a$ + Max pooling          & $87.9$ & $90.8$    \\
$H_a$ + Max pooling          & $69.7$ & $71.3$    \\
% TOADD: $H_a$ with avg. pool! since it is better than maxpool.
\hline
\end{tabular}
\end{center}
\caption{{\bf Walk aggregation methods.} 
Out attention approach outperforms other possible aggregation strategies on 3D-Future.
}
\label{tab:agg_compare}
\end{table}

% Walks length
\vspace{0.05in}
\noindent
{\bf Walk length. }
The longer the walk, the better the performance.
However, once the walk reaches $0.3$ of the vertices, the performance does not improve further.
For instance, on SHREC11, we get $90.5$ accuracy when the walk contains $0.1$ of the vertices,  $99.2$  accuracy for  $0.2$ of the vertices, and  $99.7$ accuracy for  $0.3$ or more of the vertices.

% Table~\ref{tab:length} shows the effect of the walk length on performance (for SHREC11 classification results).
% The performance improves with walk length, until the walk covers $0.3$ of the mesh vertices.
% % =================== walks length SHREC11 ============== %
% \begin{table}[h]
% \begin{center}
% \begin{tabular}{|l|c|c|c|c|c|c|}
% \hline
% Walk length  &   $0.1$ &   $0.2$   &   $0.3$   &  $0.4$ & $0.5$ & $1.0$\\
% \hline\hline
% AttWalk        &   $90.5$  &   $99.2$  &   $99.7$  &  $99.7$  & $99.7$ & $99.7$  \\
% \hline
% \end{tabular}
% \end{center}
% \caption{{\bf Walk length.} The longer the walk, the better the performance.
% The performance stabilizes when the walk reaches $0.3$ of the mesh's vertices.
% }
% \label{tab:length}
% \end{table}

\begin{figure}[h]
\centering
\begin{tabular}{cc}
\includegraphics[width=0.38\linewidth]{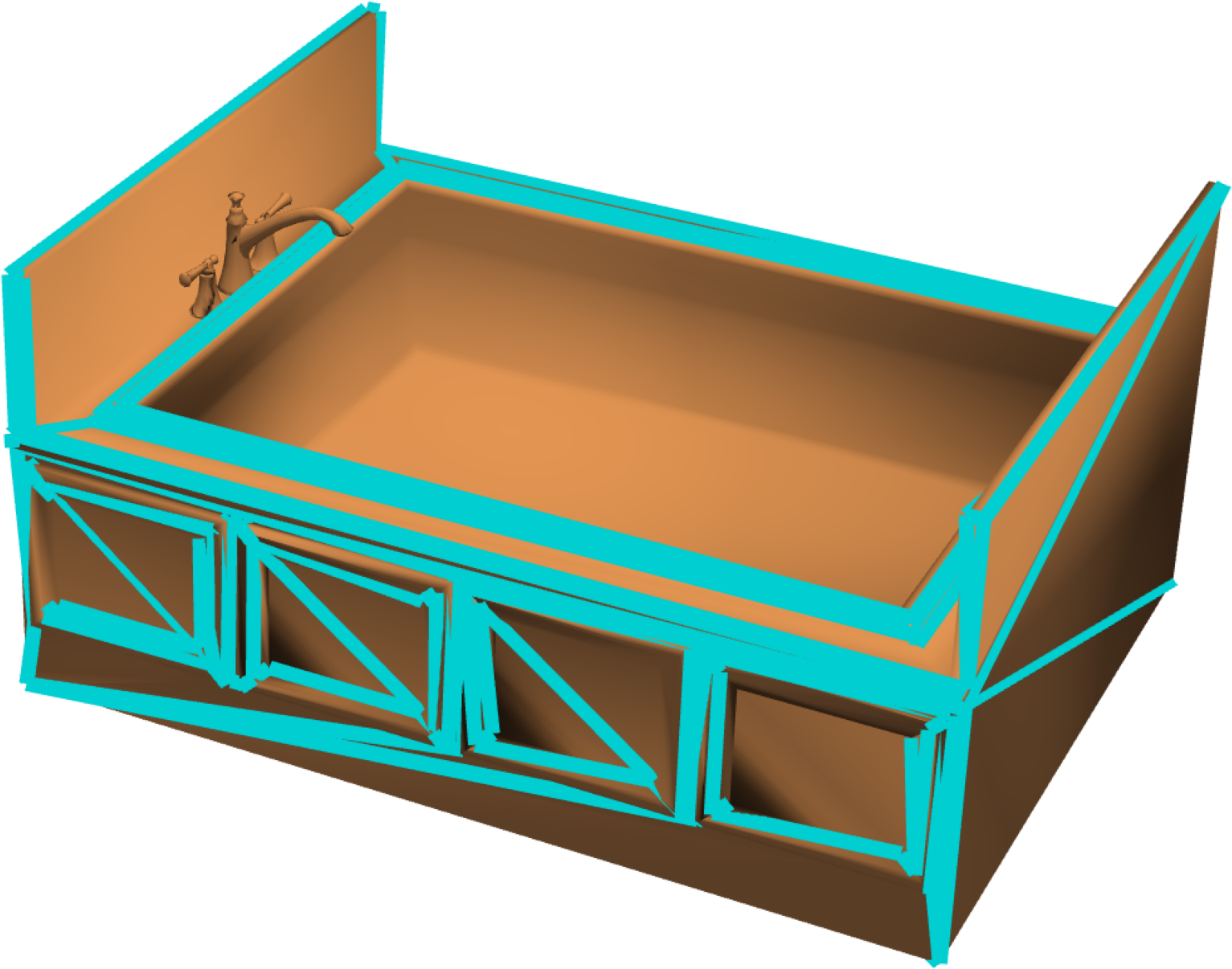} &
\includegraphics[width=.39\linewidth]{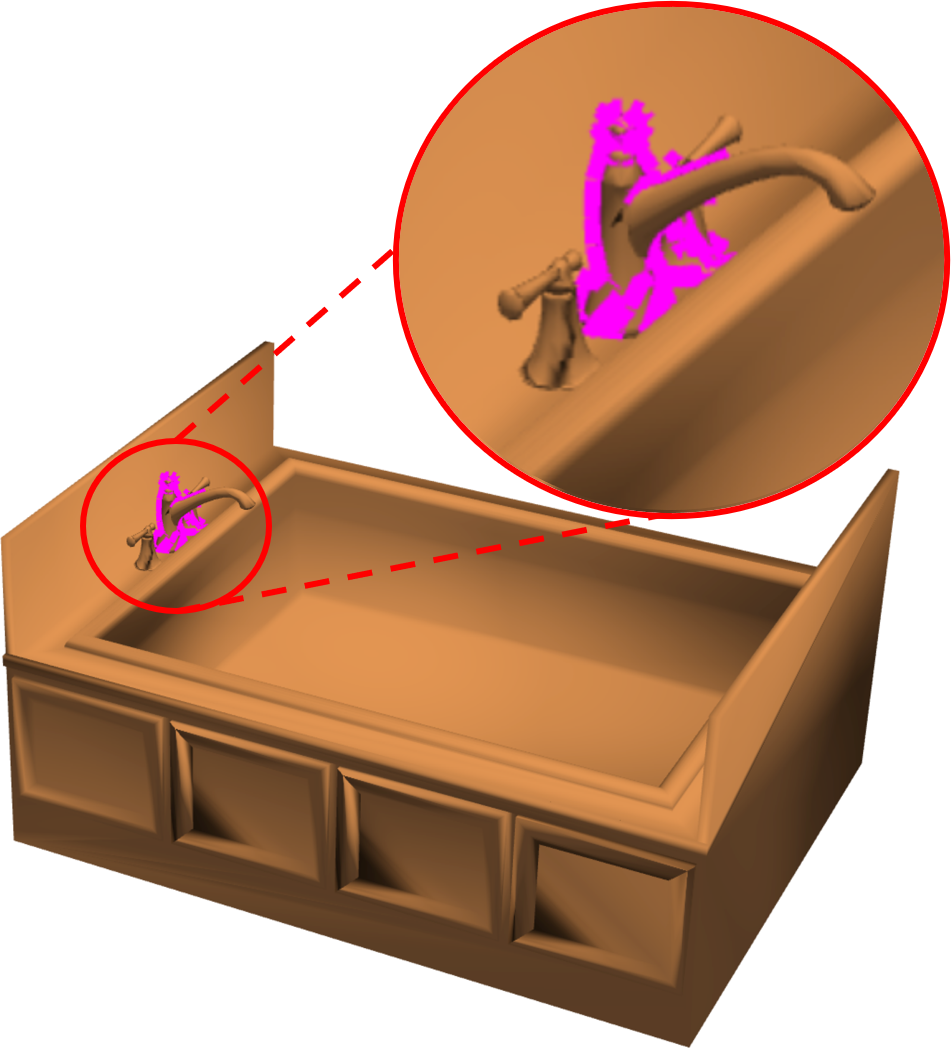}
\end{tabular}
   \caption{\textbf{Limitation.}
   Our algorithm classifies the bathtub as a bed, giving more attention to features resembling a bed, as shown by the most attentive walk (in cyan). 
   %and the least attentive walk (zoomed-in) classifies it as a sink.
%   of the bathtub causes the most attended (cyan) walk confuse it as a bed.
%   While a different walk (magenta) explore the sink handle of the bath, it is given lower attention rank, which we assume is due to its extreme locality.
   }
\label{fig:limitation}
\end{figure}

\vspace{0.05in}
\noindent
{\bf Limitations.}
Figure~\ref{fig:limitation} illustrates a failure case, where a majority rule of~\cite{lahav2020meshwalker} would be preferable.
%, where a bathtub is erroneously classified as a bed.
Though $5$ of $8$ walks indicate that the shape is a bathtub, our attention gives more weight to features that indicate that this is a bed.
The most attentive walk  (in cyan), which provides a global view of the shape, classifies it as  a bed due to the special shape of this bathtub.
The least attentive walk (in magenta) visits mostly the tap and thus classifies the bathtub as a sink.
% The most attentive walk  (in cyan), which provides a global view of the shape, classifies it as  a bed due to the special shape of this bathtub.
% The least attentive walk (in magenta) visits mostly the tap and thus classifies the bathtub as a sink.

% We fail on objects in some of the following scenarios:
% a) Objects with features which strongly resemble other classes, as the rectangular bathtub with frame which resembles a bed.
% b) Objects with features shared by multiple classes, which are at the same time crucial to the correct class classification: Taps shared by bathtub and sink classes, vases & plants shared by vase, flower pot & plant classes. 
% c) objects containing out-of-distribution details: decorations & ornamentations 

% Limitation figure %%%%%%%%%%%%%%%%%%%

%%%%%%%%%%%%%%%%%%%%%%%%%%%%
%%%% section: Conclusion
%%%%%%%%%%%%%%%%%%%%%%%%%%%%
\section{Conclusion}
This paper introduced attention into a 3D learning framework.
It showed how multiple random walks along the surface may jointly indicate the most attentive features of a 3D mesh.
The key idea is that exploring the mesh in different ways, by different walks, can be leveraged for both learning the meaningful attributes of the surface and to reduce the number of walks needed.
Our approach achieves state-of-the-art results for shape classification and shape retrieval on commonly-used datasets.

In the future, we intend to adapt our approach to other applications, most notably shape segmentation.
%and shape similarity.
Using our approach within multi-scale is also a direction that worth further studying.
%we suggest further exploring how to utilize attention in order to learn how to sample the walks as part of a deep learning system, instead of randomly walking.
%One possible way to achieve that is using reinforcement learning framework.
% %
% Another limitation is due to our cross walk attention focus on walk features, which cannot be attributed to specific vertices along the walk, making it less suitable for semantic segmentation and shape correspondence.

\newpage
{\small
\bibliographystyle{ieee_fullname}
\bibliography{main}
}

\end{document}